\def\year{2022}\relax
\documentclass[letterpaper]{article} 
\usepackage{aaai22}  
\usepackage{times}  
\usepackage{helvet} 
\usepackage{courier}  
\usepackage[hyphens]{url}  
\usepackage{graphicx} 
\usepackage{natbib}  
\usepackage{caption} 
\usepackage{epsfig}
\usepackage{amsmath}
\usepackage{amssymb}

\newcommand{\eg}{{\em e.g.}}
\newcommand{\ie}{{\em i.e.}}
\newcommand{\etal}{{\em et al.}}

\usepackage{xcolor}

\usepackage{scalerel}
\usepackage{array}
\usepackage[linesnumbered,ruled,vlined]{algorithm2e}
\usepackage{subcaption}
\usepackage{mathtools}
\usepackage{dblfloatfix}

\usepackage{tcolorbox}
\newtcolorbox{mybox}[3][]
{
  colframe = #2!50,
  colback  = #2!10,
  coltitle = #2!10!black,  
  title    = {#3},
  boxsep   = 0.25pt,
  left     = 0.5pt,
  right    = 0.5pt,
  top      = 0pt,
  bottom   = 0pt,
  width=\linewidth,
  #1,
}

\makeatletter
\newcommand{\removelatexerror}{\let\@latex@error\@gobble}
\makeatother 

\newcommand{\mcircle}{$\mathbin{\scalerel*{\circ}{j}}\,$}
\newcommand{\mtriangle}{$\Delta\,$}
\newcommand{\mplus}{$+\,$}
\newcommand{\tr}[2][]{\ensuremath{#2^{\text{tr}}_{#1}\,}}
\newcommand{\te}[2][]{\ensuremath{#2^{\text{te}}_{#1}\,}} 

\nocopyright
\title{Ordinal-Quadruplet: Retrieval of Missing Classes in Ordinal Time Series}
\author {
    Jurijs Nazarovs,\textsuperscript{\rm 1}
    Cristian Lumezanu, \textsuperscript{\rm 2}
    Qianying Ren, \textsuperscript{\rm 3} 
    Yuncong Chen, \textsuperscript{\rm 2} 
    Takehiko Mizoguchi,\textsuperscript{\rm 2}
    Dongjin Song,\textsuperscript{\rm 3}
    Haifeng Chen\textsuperscript{\rm 2}\\
}
\affiliations {
    \textsuperscript{\rm 1} {University of Wisconsin, Madison} \\
    \textsuperscript{\rm 2} {NEC Laboratories America} \\
    \textsuperscript{\rm 3}{University of Connecticut}\\
    nazarovs@wisc.edu, lume@nec-labs.com, yuncong@nec-labs.com,tmizoguchi@nec-labs.com,dongjin.song@uconn.edu,haifeng@nec-labs.com
}

\begin{document}

\maketitle

\begin{abstract}
In this paper, we propose an ordered time series classification framework that is robust against missing classes in the training data, \textit{i.e.}, during testing we can prescribe classes that are missing during training. 
This framework relies on two main components: (1) our newly proposed ordinal-quadruplet loss, which forces the model to learn latent representation while preserving the ordinal relation among labels, (2) testing procedure, which utilizes the property of latent representation (order preservation). We conduct experiments based on real world multivariate time series data and show the significant improvement in the prediction of missing labels even with 40\% of the classes are missing from training. Compared with the well-known triplet loss optimization augmented with interpolation for missing information, in some cases, we nearly double the accuracy.
\end{abstract}

\section{Introduction}
{\em Ordinal time series} is data collected over time, from one or more sources (or sensors), and whose labels form an order relationship. For example, the progression of operational states in a power plant, \eg{}, shutdown, bootstrap, partial load, and full load, forms an ordered set of labels. Similarly, the labels assigned to air quality index measurements over time (\eg{}, good, moderate, sensitive, unhealthy, very unhealthy, hazardous~\cite{sowlat2011novel}) are ordered. 
Increasingly, many real world applications and systems, such as power plants, networks, health and patient monitoring platforms, IoT ecosystems, or fitness devices, generate ordinal time series.
Interpreting and classifying the current status of these systems is critical for their uninterrupted operation.

In the past, many different techniques have been developed to analyze time series data without explicitly considering its label order, \textit{e.g.},  RNN~\cite{graves2013speech}, LSTM~\cite{deep-rth}, TDCNN~\cite{bai2018empirical}, DA-RNN~\cite{qin2017dual}), and autoregressive models~\cite{akaike1969fitting,mcleod1983diagnostic}). To classify new time series data, 
these methods looks for a compact representation of the raw data
for retrieval~\cite{weinberger2006distance} or regression to interpret the status from historical data~\cite{wang2017deep}. Critically, they all assume independence among labels. On the other hand, the key aspect of ordinal classification is that \textit{not all wrong classes are equally wrong}~\cite{diaz2019soft}. Knowing an order among classes could improve classification by filtering out predictions that are too {\em far away} to be correct. Recent work on non-sequence data uses ordinal relationships among labels to improve classification accuracy for age detection tasks~\cite{chang2011ordinal}.

Another potential benefit of ordinal classification, which \textbf{we explore in this paper}, is the ability to work with incomplete training data. Indeed, if we are able to capture the order relationship among the labels, we may afford to have {\em gaps} in the training data (\eg{}, data for some classes completely missing or only represented by a few samples) without losing much accuracy. Understanding and controlling the trade-off between the amount of missing data and accuracy would give users more flexibility in improving the overall performance of a classifier.

\begin{figure*}[t]
\centering
\begin{subfigure}{0.32\textwidth}
\centering
\includegraphics[height=0.16\textheight]{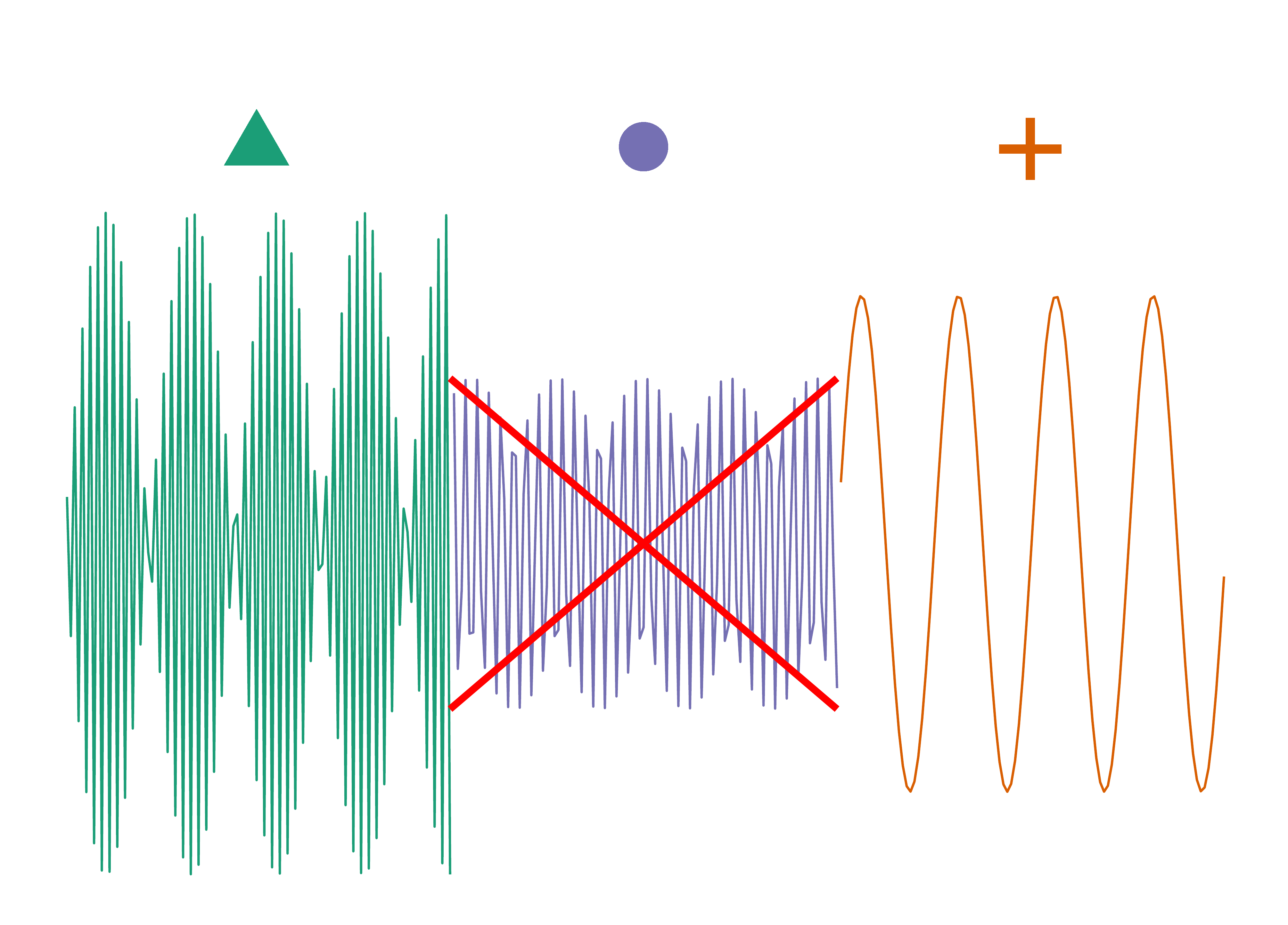} 

\caption{Time series segments with ordinal labels: \mtriangle$<$\mcircle $<$\mplus. During training, data with class \mcircle is omitted. \label{fig:timeseries}}
\end{subfigure}\hfill
\begin{subfigure}{0.32\textwidth}
\centering
\includegraphics[height=0.16\textheight]{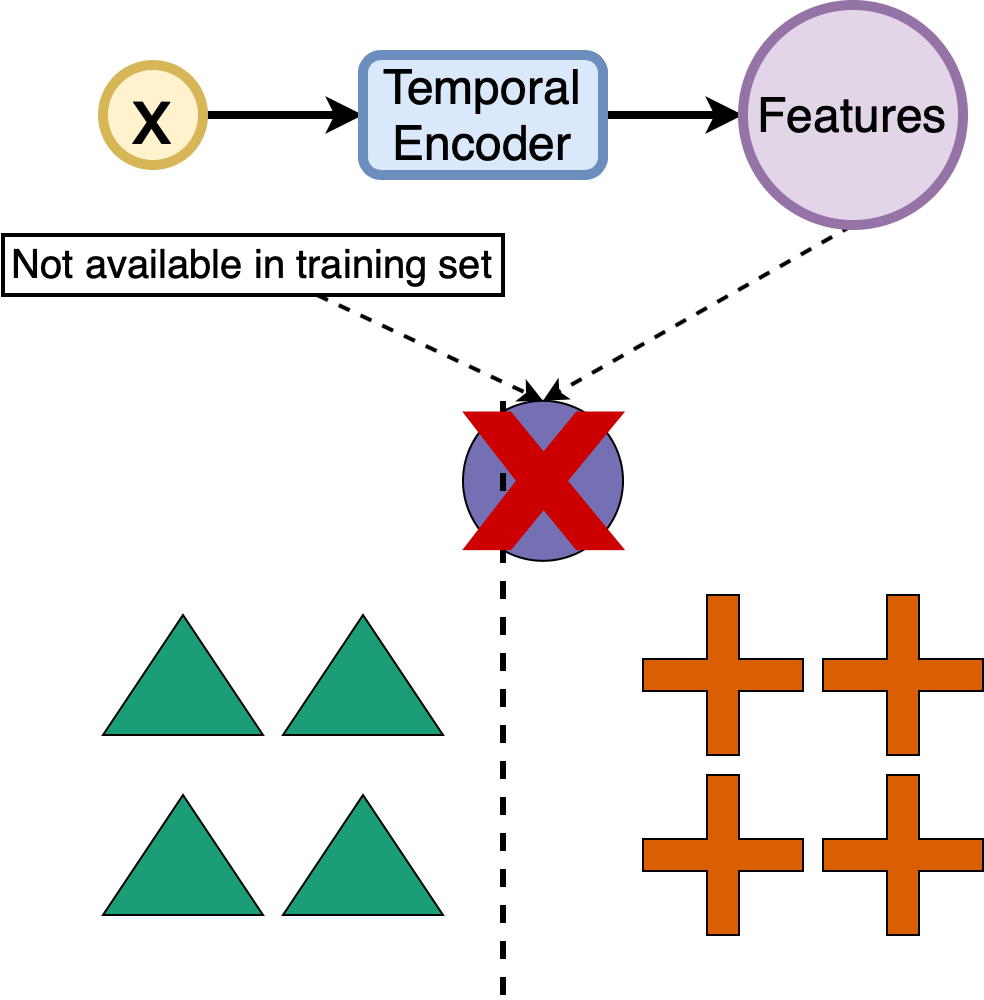} 
\caption{k-nn prediction with a triplet loss. Absence of class \mcircle in training data results in biased prediction, assigning either \mtriangle or \mplus . \label{fig:knn}}
\end{subfigure}\hfill
\begin{subfigure}{0.32\textwidth}
\centering
\includegraphics[height=0.16\textheight]{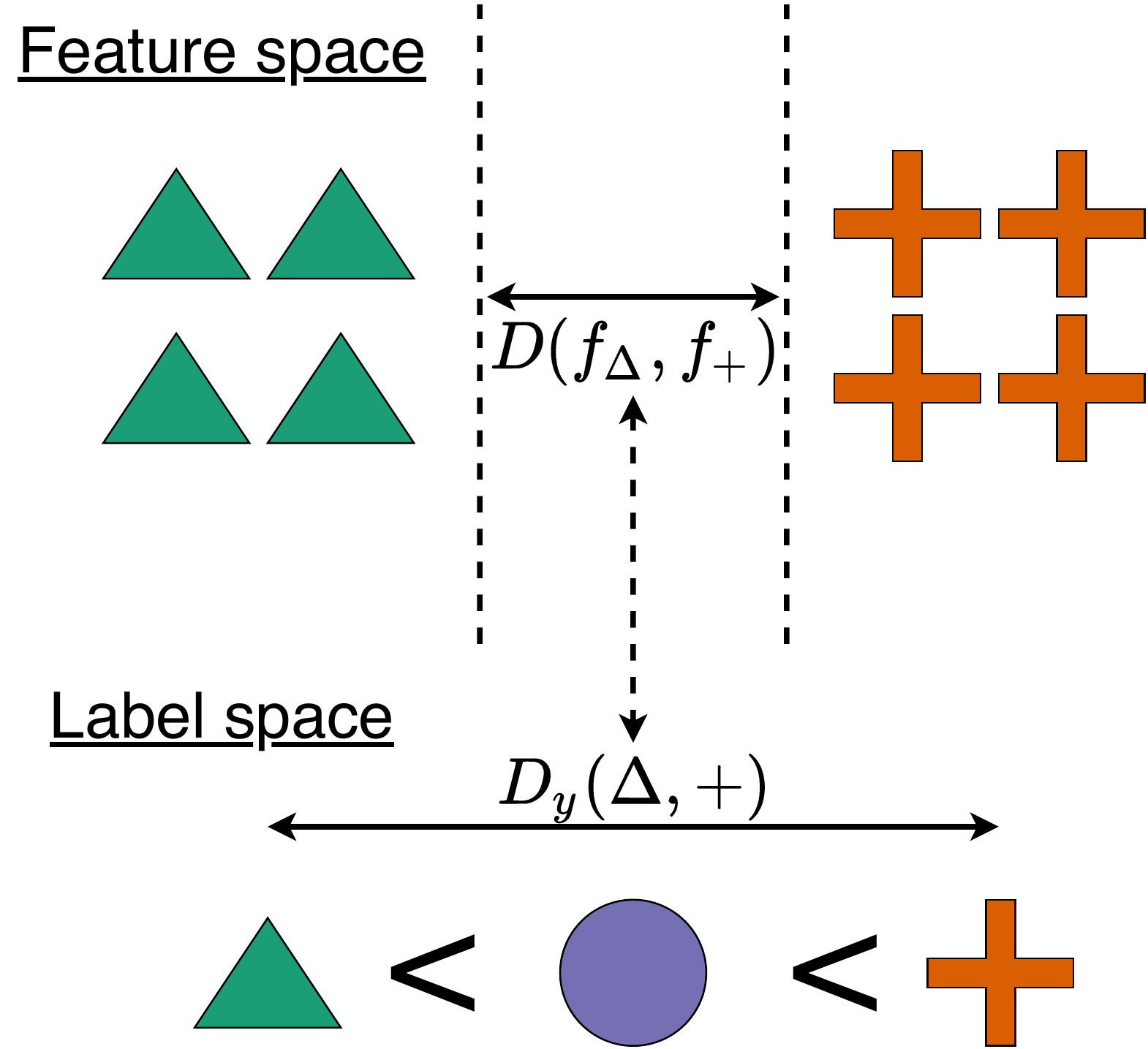} 
\caption{By preserving same order of distance between feature space  and label space, we acknowledge the existence of missing classes. \label{fig:ordinal_dist}}
\end{subfigure}
\caption{Ordinal relation in time series data}
\end{figure*}

In this paper, we propose a time series ordinal classification framework, 
which at a testing time can prescribe the labels of classes that are omitting during training. 
Our framework 
consists of a
temporal encoder to map segments of multivariate time-series in latent space,
a custom loss function that preserves ordinal relation of labels in the embedded space,
and hypothesis test to be able to classify test samples with labels not present in the training data. Using several time series datasets with ordinal relation in classes, we show that our framework maintains overall prediction accuracy even when up to 40\% of all possible labels are completely missing during training.

We bring the following \textbf{contributions}. First, we introduce a novel loss function, called {\em ordinal-quadruplet}, to capture the order between labels and encode it into the latent space of time series. Ordinal-quadruplet relies on (1) a triplet component~\cite{hermans2017defense} to separate embeddings of time series with different labels and bring closer embeddings from the same label, and (2) on a log ratio component~\cite{kim2019deep} to keep the distance between embeddings proportional to the distance between their associated labels. Second, given latent space with preserved order of labels, achieved by ordinal-quadruplet, we propose a method to classify new time series samples even when they belong to classes not present in the training data. Our method is built on 
finding the ``closest'' class based on the ranking similarity between label space and learned latent space.
\section{Background and goals}
\subsection{Ordinal classification}
Recent years have witnessed an increased interest in classification (also called ordinal regression) of ordinal data using neural networks. There are several ways to capture the ordinal information and reflect it into the network parameters and output.

First, encoding order into the optimization {\bf loss} helps compute embeddings that preserve label ordering. Kim~\etal\cite{kim2019deep} propose the log-ratio loss, which maintains the same distance ratio between embeddings and their associated labels.
Despite the ability to preserve the ratio of distances and thus the order of labels, the log-ratio loss does not maintain class similarity (\ie{}, samples from the same class should have similar embeddings). In our framework, we devise a novel loss function that combines the log-ratio loss with a similarity preserving loss function such as the triplet loss.

{\bf Sampling} good data for optimization is an essential part of any learning framework.
As noted in \cite{hermans2017defense}, learning with triplet loss performs better with hard triplets, where the negative sample is closer to the anchor than positive. Wu ~\etal \cite{wu2019ordinal} propose to sample triplets $(a, s, d)$ whose labels satisfy an order relationship, \ie{}, $|y_a - y_d| > |y_a - y_s| + \alpha$, where $\alpha$ is a preset parameter. However, incorporating this constraint in sampling may lead to class imbalance where there are much more samples from the middle of the order than the ends.

{\bf Soft labels} incorporate order-violationg penalties into the label representation rather than the network optimization. A soft label is a vector associated with each normal data label and which represents the distance from the label to all other labels~\cite{tzeng2015simultaneous,tan2016age,diaz2019soft}. It is an intuitive way of embedding ordinal information and works with off-the-shelf deep networks, without any modifications to the learning pipeline. However, soft labels require the network optimization loss to fit the soft form of labels (\eg{}, cross-entropy or Kullback-Leibler) and does not work for retrieval-based classification frameworks that use distance between network embeddings to identify the correct label~\cite{deep-rth}. 

\subsection{Missing data imputation}
Missing information is very common problem in broad areas of research \cite{kang2013prevention} and occurs in both covariates and response. To deal with incomplete data in coveriates, machine learning community provides methods based on k-nn \cite{batista2003study}, neural networks \cite{sharpe1995dealing, smieja2018processing}, and iterative techniques \cite{buuren2010mice}. However, not much attention is given to problems with \textbf{missing classes or labels}, especially in high dimensional temporal settings. One of the examples is human activity identification based on sensors, which generate 100s attributes. During experiments it is possible that subject can perform just part of activities from the domain, either because of health issues, time constraints or budget limits. Thus, it is important to have a model, which can predict the class of the testing sample, even if no data of this class was involved in training.

\subsection{Problem statement and goals}
\label{subsec:statement}

\textbf{Usual time series classification.}
Consider a multivariate time series classification problem with more than three classes. A common approach for classification is to learn latent representation for time series segments by applying a temporal encoder that optimizes a metric loss function (\eg{}, the triplet loss). The encoder projects the time series into a feature (or embedding) space $f(\cdot)$, where similar data is clustered together, while dissimilar segments are far apart. 
To classify new test data \te{x}, we extract its representation $f(\te{x})$ and predict its class using the k-nearest neighbors (k-nn) based on the embedding of the training data $f(\cdot)$.

\textbf{Ordinal classification with missing labels during training.}
Usually ordinal classification papers focus on incorporating ordinal information for accuracy improvement, given that training and testing data contain same classes. However, it is important to understand the benefits of such integration on post-training inference, when training data does not cover all classes, which can be observed during testing.

Unlike usual time series classification problems, we have two additional assumptions: (ordering) there is a total ordering relationship among the classes and (possible missing labels) some classes may not have any samples in the training data.
To understand the challenge, consider the motivation example, when we focus only on three classes, with labels:  \mtriangle, \mcircle, and \mplus, such that \mtriangle $<$ \mcircle $<$ \mplus, Figure~\ref{fig:timeseries}. 
Assume that during training, data corresponding to \mcircle is not available, and we learn to discriminate features in latent space based on two classes: \mtriangle and \mplus. At test time, a use of techniques like k-nn leads to a biased prediction (Figure~\ref{fig:knn}): even if the true label of a testing sample is \mcircle, we are able to predict only \mtriangle or \mplus.

A solution, when knowing the complete set of labels (\mtriangle, \mcircle, and \mplus) and their order (\mcircle is between \mtriangle and \mplus), is to interpolate 
the center of a missing class in the feature space $f(\cdot)$ as a middle point between the centers of its neighbors: $\bar{f}_{\bigcirc} := \frac{1}{2}\bar{f}_{\Delta} + \frac{1}{2}\bar{f}_{+}$, where 
$\bar{f}_{\Delta}, \bar{f}_{+}$ denote the average of features for samples with labels $\Delta$ and $+$.
For new data, we assign the class with the closest center: $\bar{f}_{\bigcirc}, \bar{f}_{\Delta}$, or $\bar{f}_{+}$.

However, there are several issues with interpolation.
First, it assumes that the center of the features of missing class $\bar{f}_{\bigcirc}$, is in the middle of the interval between centers of features for known labels (\mtriangle, \mplus): $\bar{f}_{\Delta}$ and $\bar{f}_{+}$.
 
Second, assigning labels based on the center of features does not account for the variation of features for each label (\eg{}, some classes have samples with features spread apart, others close together).
Third, if the boundary class (e.g. \mtriangle or a \mplus, in our case) or if multiple consecutive classes are missing, it is not clear how to interpolate.  

Although three aforementioned issues are important and have to be addressed, one of the main assumptions is that relative order of classes is preserved in the feature space $f(\cdot)$. While it is true in case when there are only three possible classes (\mtriangle, \mcircle, and \mplus) with only one being missing from training (\mcircle), either $\bar{f}_{\Delta} < \bar{f}_{+}$ or $\bar{f}_{+} < \bar{f}_{\Delta}$, it is not true in general with more classes. Thus, we seek to learn a latent space that preserves information about the label order.
We do this by ensuring that distances between learned features are proportional to distance between their associated labels (see Figure~\ref{fig:ordinal_dist}). Such regularization implicitly limits the distance in the feature space from missing classes to its neighbors and gives us a search radius to perform post-training inference.

\section{Our method}
\label{sec:method}

\begin{figure}[ht]
\centering
\includegraphics[width=\columnwidth]{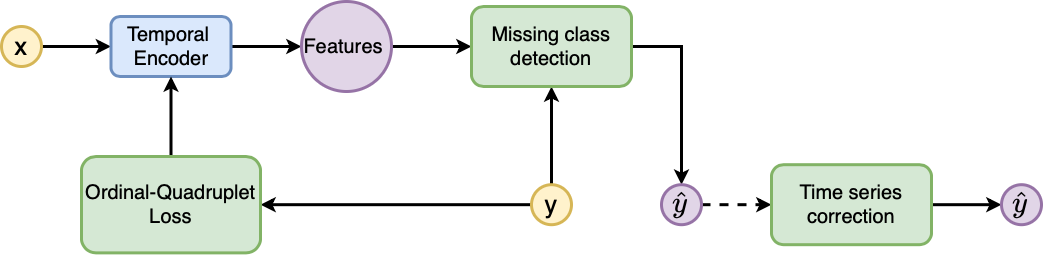} 
\caption{Ordinal-quadruplet pipeline for missing classes recovery. Our contribution: loss to preserve distances; statistical based module to recover missing classes, based on order similarity; correction of prediction based on time series assumption. \label{fig:pipeline}}
\end{figure}

In this paper we propose a framework for ordinal classification of time series, which enables imputation of missing labels with high accuracy. The framework consists of two parts (Figure~\ref{fig:pipeline}):
\textbf{first}, we propose a new ordinal-quadruplet loss to force temporal encoder to learn latent space preserving ordinal relation between classes similar to label space;
\textbf{second}, we present an ordinal retrieval-based classification module to make prediction for missing classes.
\textit{Note} that the choice of temporal encoder is not important for our method, as long as it maps time series segment to latent representation.
Through this section we use notation defined in Table~\ref{tab:notation}.

\begin{table}[ht]
\begin{tabular}{ |c|m{6cm}|}
 \hline
 $S$ & domain of all classes\\
 $N$ & set of non-missing classes \\
 $M$ & set of missing classes\\
 \tr{x}, \te{x} & training and testing sample, respectively\\
 $y(x)$ & label of sample $x$\\
 $f(x)$ & learned features of sample $x$\\
 $\overline{f_c}$ & mean $\{f(\tr[i]{x}): \forall i, y(\tr[i]{x})=c\}$\\
 \hline
 $D(f_i, f_j)$ & distance between 2 features: $\left\|f_i - f_j\right\|_2^2$\\
 $D_y(y_i, y_j)$ & distance defining labels ordinal relation\\
 $\mathbf{d_y}$ & $\{D(f(\tr[i]{x}), \overline{f_y}): \forall i, y(\tr[i]{x})=y\}$\\
 $\te[y]{d}$ & $ D(f(\te{x}), \overline{f_y})$\\
 $Q(p, d)$ & $p$-th sample quantile of the set $d$\\
  \hline
\end{tabular}
\caption{Notation used through the Section~\ref{sec:method}\label{tab:notation}}
\end{table}

\subsection{Preserving order: Ordinal-quadruplet loss}

\begin{figure*}[ht!]
\centering

\begin{subfigure}{0.8\textwidth}
\centering
\includegraphics[trim={4.4cm 7.3cm 3.4cm 30cm}, clip,
height=0.03\textheight]{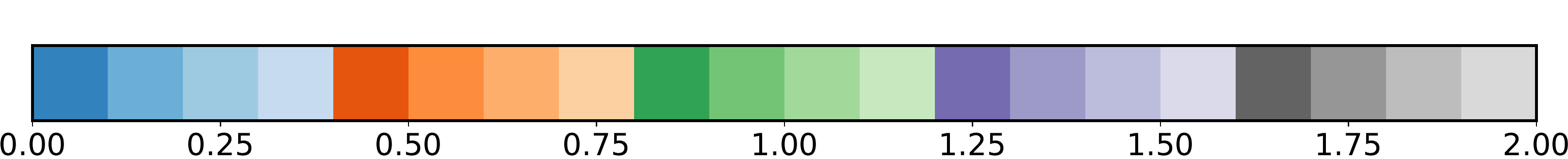} 
\end{subfigure}

\begin{subfigure}{0.33\textwidth}
\centering
\includegraphics[trim={4.2cm 5cm 4.2cm 3.6cm}, clip, width=0.6\textwidth]{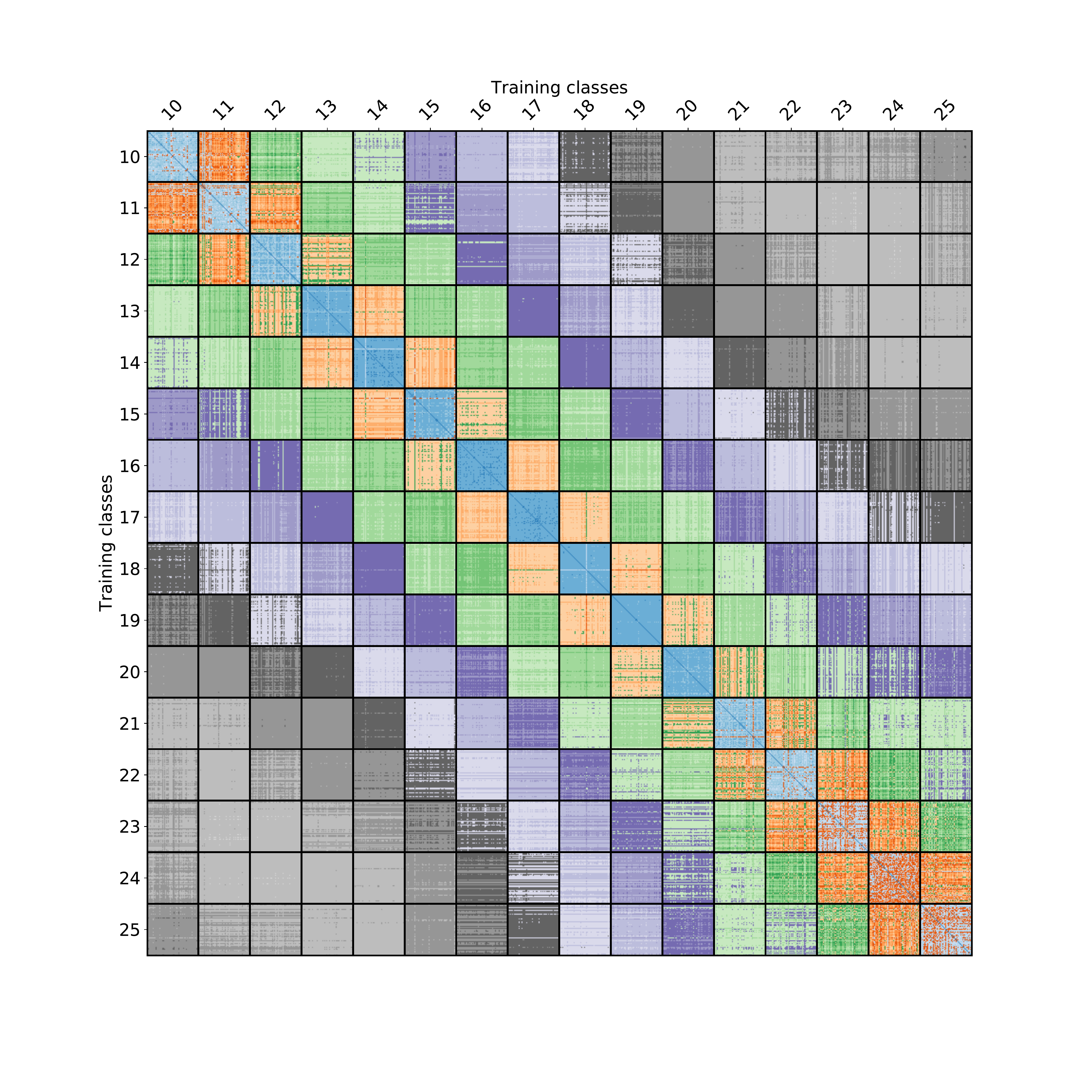} 
\caption{OQ with $D_y = |y_i-y_j|$}
\end{subfigure}
\begin{subfigure}{0.33\textwidth}
\centering
\includegraphics[trim={4.2cm 5cm 4.2cm 3.6cm}, clip, width=0.6\textwidth]{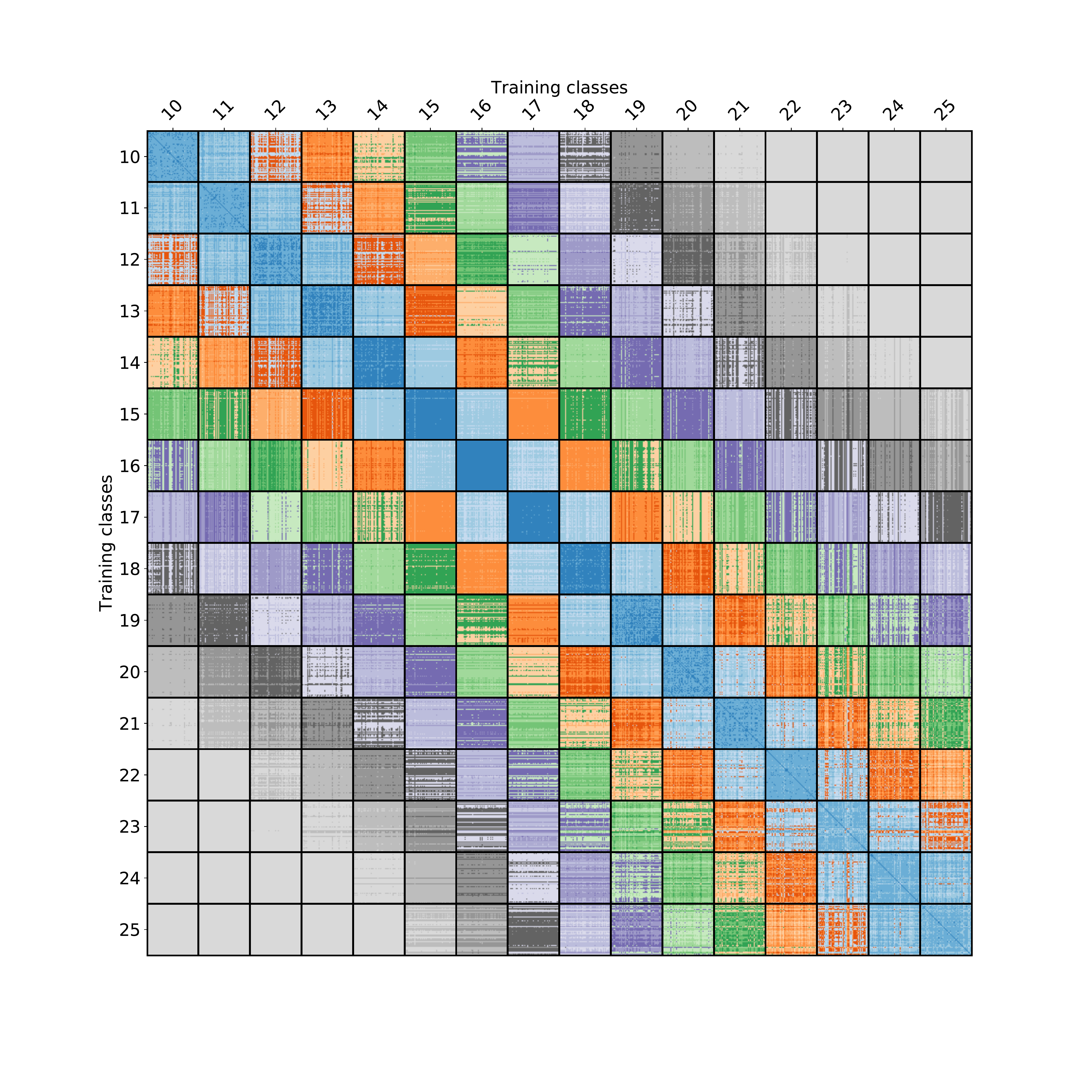} 
\caption{OQ with $D_y = (y_i-y_j)^2$ }
\end{subfigure}
\begin{subfigure}{0.33\textwidth}
\centering
\includegraphics[trim={4.2cm 5cm 4.2cm 3.6cm}, clip, width=0.6\textwidth]{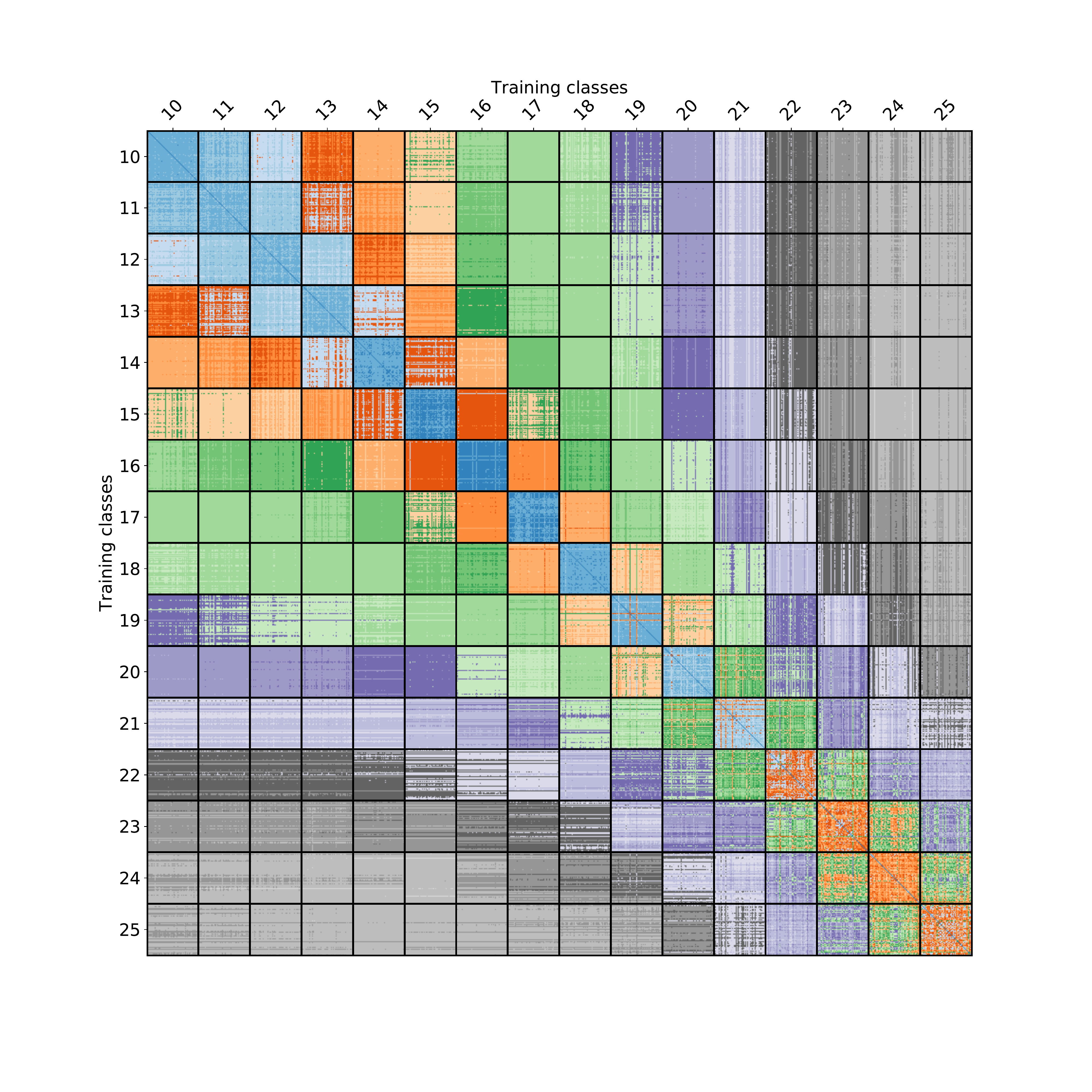} 
\caption{OQ with $D_y = |10^{\frac{y_i}{10}} - 10^{\frac{y_j}{10}}|$}
\end{subfigure}

\caption{Pairwise distances in feature space of the trained encoder between samples of different classes, based on $D_y$. \label{fig:different_dy}}
\end{figure*}

Our first task is to build a time series encoder that learns latent representation of the data to discriminate data with different labels, meanwhile preserving the order of labels. We have the following requirements:
\begin{enumerate}
    \item small intra-label distances, \ie{}, samples with the same label should have similar features
    \item large inter-label distances, \ie{}, samples with different labels should have features that are far apart
    \item ordinal relation, \ie{}, distances between features of samples with different labels preserves the order of their labels.
\end{enumerate}

Various existing metric loss based optimizations can achieve some of the requirements above, but to the best of our knowledge no existing optimization can achieve all at the same time. Triplet loss~\cite{kumar2016learning} selects triplets of samples such that two samples are of the same label and the third is a different label. They can help discriminate among classes, but do not preserve order, as only two labels are involved at each step of the optimization. The log-ratio loss~\cite{kim2019deep} encodes the order by selecting triplets from three different labels. However, it does not explicitly reduce the intra-label distance, since no samples have the same label. 

To address all three requirements at the same time, we propose the {\bf ordinal-quadruplet loss}, where we select a quadruplet, rather than a triplet, ensuring that two of the samples are of the same label, while the other two are of different labels from the first samples and from each other.
For a quadruplet sample $(a, s, i, j)$, with labels $y$, such that $y_a = y_s$ and $y_i \neq y_a,  y_j \neq y_a$, and learned latent representation $f$, \textbf{ordinal-quadruplet loss} is defined as:
$$
L = l_t(a, s, i) + l_t(a, s, j) + l_{lr}(a, i, j), \text{ where}
$$
$$
l_t(a, p, n)=\left[D\left(f_{a}, f_{p}\right)-D\left(f_{a}, f_{n}\right)+\delta\right]_{+},
$$
$$
l_{\mathrm{lr}}(a, i, j)=\left\{\log \frac{D\left(f_{a}, f_{i}\right)}{D\left(f_{a}, f_{j}\right)}-\log \frac{D_y\left(y_{a}, y_{i}\right)}{D_y\left(y_{a}, y_{j}\right)}\right\}^{2}
$$
Ordinal relation is defined by $D_y(\cdot, \cdot)$ and can be as simple as $D_y(y_i, y_j) = |y_i - y_j|$ or based on prior knowledge about the data. 
In Figure~\ref{fig:different_dy} we demonstrate different choices of $D_y(\cdot, \cdot)$ and how it effects the learned space. 
Next, we emphasize the importance of maintaining the ordinal relation in the feature space.

\subsection{Detecting missing classes}
Given our ordinal-quadruplet loss, a temporal encoder learns features that preserve the ordinal relation among labels (classes), \ie{} for each label, the ranking of distances between members of different classes in latent space is similar to ranking of distances in label space. Having this property brings a different perspective on classification using latent space.

\subsubsection{Distance rank correlation.} 
\begin{figure}[ht]
\centering
\includegraphics[width=0.8\columnwidth]{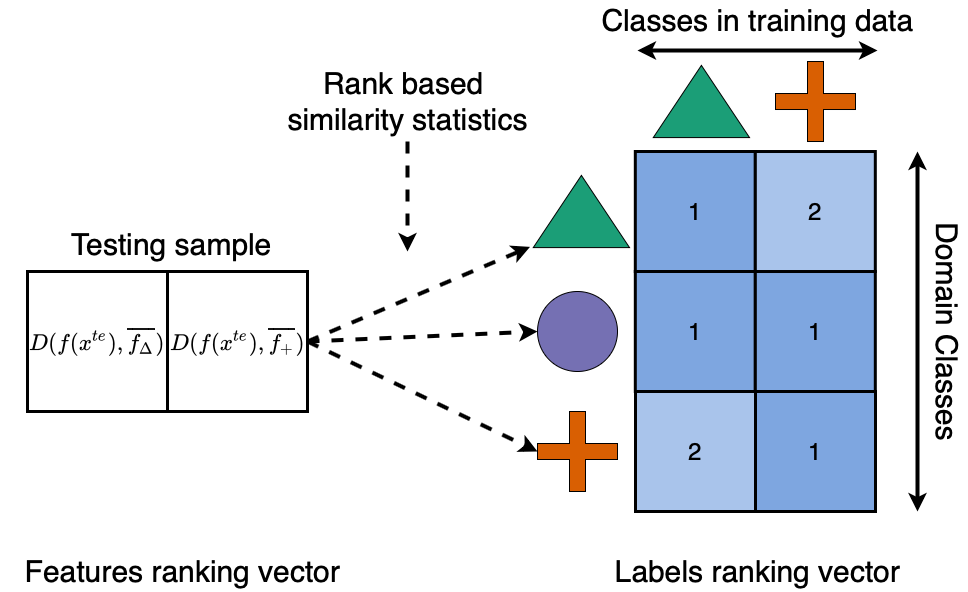} 
\caption{1) Left: in feature space, for a testing sample \te{x} compute distances to center of corresponding clusters for every class presented in training data. Compute rank of these distances and denote it as ``features ranking vector''. 2) Right: in label space, for every class in the domain (rows), compute distance $D_y(\cdot, \cdot)$ to every class presented in training data (columns). Compute rank of these distances and denote it as ``labels ranking vector''. Numbers in the heat-map correspond to rank of distances and not distances itself. 3) Use rank based similarity statistics between ``features ranking vector'' (Left) and `` labels ranking vector'' (Right) to prescribe a class to \te{x}. \label{fig: retrieval}}
\end{figure}

To classify new samples, we propose to use rank-based statistics, such as Spearman's $\rho$, Kendall's $\tau$, Goodman and Kruskal's $\gamma$, and Somers' $D$ (see \cite{goktas2011comparison} for more information), instead of the more popular k-nn based retrieval. The advantage of rank-based statistics over k-nn retrieval is 
an ability to classify a testing sample as a missing (from the training data) class. 
For a testing sample, it can be done by finding the ``closest" class 
based on relation between distances in label space and distances in feature space. The ``closest'' is defined by one of rank-based statistics, mentioned above. To have a better connection with the motivation example, we describe our method in caption of Figure~\ref{fig: retrieval}.

However, simply selecting the ``closest'' class (using ranking based statistics) as a prediction does not work when there is a tie. Consider Figure~\ref{fig: retrieval} (right), our testing sample's true class is \mcircle which is missing from the training data. The label ranking vector of \mcircle is $(1,1)$. If our encoder perfectly matches relation between labels into the feature space, \ie{}, $D(f(\te{x}), \overline{f_\Delta}) = D(f(\te{x}), \overline{f_+})$, this would not be an issue as $(1,1)$ would be the ``closest'' label ranking vector to the ranking vector of distances in latent space. However, in practice, the feature distances are rarely equal, \ie{}, either $D(f(\te{x}), \overline{f_\Delta}) > D(f(\te{x}), \overline{f_+})$ or opposite, which results in labels ranking vector be either $(2, 1)$ or $(1, 2)$. This, in turn will incorrectly predict classes \mtriangle or \mplus for our testing sample.

However, if order of labels is almost preserved in latent space, in case the ``closest'' class is a wrong one and does not correspond to a missing class, the second ``closest'' class is the right one. Based on this observation we propose the following algorithm to assign class to the testing sample, which is capable to assign classes missing from the training data (follow notation in Table~\ref{tab:notation}):

\begin{mybox}{gray}{}
\removelatexerror
\begin{algorithm}[H]
    \KwOut{Predicted label for \te{x}}
    Compute pairwise distances between domain classes and non-missing classes,  Figure \ref{fig: retrieval} (right): $L = \{D_y(s, n): \forall s \in S, \forall n \in N\}$, size: $|S|\times|N|$
    
     For a testing sample \te{x}, compute distances to $\overline{f_n}$ for every class $n \in N$, Figure \ref{fig: retrieval} (left):
    $F = \{\te[n]{d}: \forall n \in N\}$, size: $1\times|N|$
    
     For every class $s\in S$ compute rank-based statistics between $F$ and $L_s$ (row of the matrix $L$ corresponding to class $s$). Select two classes $s_1, s_2 \in S$ with highest value of the statistics
 
     \uIf {$s_1, s_2 \in N$}{predict class using k-nn}{
        \uElseIf{$s_1, s_2 \in M$}{chose class with highest rank-based statistics}{
            \Else{use our introduced below hypothesis test to decide which class (missing or non-missing) fits the best}
        }
     }
\caption{Ordinal classification with missing classes}
\label{algo:pred}
\end{algorithm}
\end{mybox}

\begin{figure}[t]
\centering
\includegraphics[width=\columnwidth]{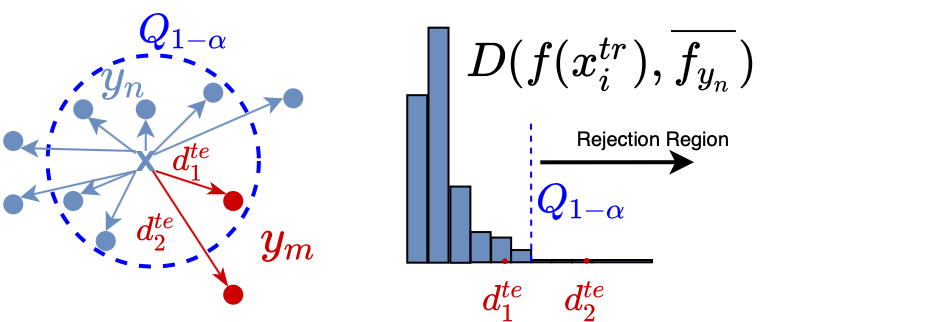} 
\caption{Intuition behind the hypothesis test. Blue circles are \tr{x} with non-missing class $y_n$, and red circles are \te{x}. \label{fig:test}}
\end{figure}

\subsubsection{Hypothesis test.}
According to Algorithm \ref{algo:pred} (last ``else''), if one of the ``closest'' classes is in set of missing (from training data) classes, and another is not, then we need a new procedure to make a decision. 
Consider the case when the two most similar classes (according to rank-based statistics) for a testing sample \te{x} are $y_n$ (present in the training data), and $y_m$ (missing from the training data). We have to decide which class is more likely to be a true label for \te{x}, given that the feature space $f(\cdot)$ does not have information about the cluster center $\overline{f_{y_m}}$ for the missing class $y_m$.

We frame the problem as outlier detection. Given learned features of training data $f(\tr[i]{x})$ with label $y_n$, we identify whether $f(\te{x})$ is an outlier and should belong to class $y_m$. Neural network encoders usually generate high dimensional feature spaces with no explicit assumptions on the distribution of the data, making outlier detection on the feature space very sensitive. Common approaches perform dimensionality reduction~\cite{kamalov2020outlier, aggarwal2001outlier} or conduct distance-based approach \cite{radovanovic2014reverse, angiulli2005distance}. Following the motivation behind distance-based approaches, we propose a non-parametric hypothesis test. The intuition behind the test is that, if testing sample \te{x} belongs to the missing class $y_m$, \te{x} should be far enough from the center of features for class $y_n$, compared to features $f(\tr[i]{x})$ of class $y_n$. Figure~\ref{fig:test} presents a visual explanation and the following definition formalizes the test:
\begin{mybox}{gray}{}
\begin{align*}
H_0: &\text{ test sample \te{x} belongs to non-missing class } y_n\\
H_A: &\text{ test sample \te{x} belongs to missing class } y_m\\
\text{RR}:  &\text{ reject } H_0 \text{ if } \te[y_n]{d} > Q(1-\alpha, \mathbf{d_{y_n}}), \text{ where }\\
&\alpha \in [0, 1] \text{ is a parameter}
\end{align*}
\end{mybox}

Our test satisfies the main definition of the hypothesis test and controls
$P(\text{Type 1 error}) = P(\text{reject } H_0 | H_0) = \alpha$. The question is how to chose $\alpha$ to compute a threshold for the rejection region $Q(1-\alpha)$. Since $P(\text{Type 1 error})=\alpha$, it can be interpreted as the percentage of data with labels $y_n$ that we can sacrifice to recover all missing labels. Commonly suggested values for $\alpha$ are 0.01, 0.05 and 0.1. In the experiments section we analyze the power of the test, \ie{}, $P(\text{reject } H_0 | H_A)$ depending on number of missing classes in the training set.

\subsection{Historical window correction}
So far our methodology can be applied to any ordinal classification problem regardless of being time series. However, the common assumption for temporal data is that the labels do not change abruptly and remain constant for a longer window of time. For example, a person keeps the same state for a period of time, e.g. sitting, or running; a power plant does not rapidly and continually shift between partial load and full load; train noise lasts longer than 0.5 seconds. Given this assumption we can improve the accuracy of missing lables prediciton.

From the nature of hypothesis test, probability of type I error is bounded by $\alpha$, which means that on average we will prescribe $\alpha$\% of wrong classes, resulting in $\alpha$\% of false positive. 
Given a window of size $w$ over which the class of the time series does not change, we can perform a real-time prediction after each sample in the window and attempt a correction at the end of the window. One possible correction method is {\em majority rule} where we first collect the most predicted class in the window and assign it to all samples in the window. Historical window correction can improve the accuracy when the time series does not change its label often or when we know when labels change~\cite{aminikhanghahi2017survey}. This comes at the expense of a delay in prediction: we have to wait until the end of a window for the final prediction for each sample. In experiments, we show that even a small window leads to significant improvement in accuracy.

\begin{figure*}[ht!]
\centering

\begin{subfigure}{0.48\textwidth}
\centering
\includegraphics[trim={1cm 2cm 1cm 9.5cm}, clip, height=0.18\textheight]{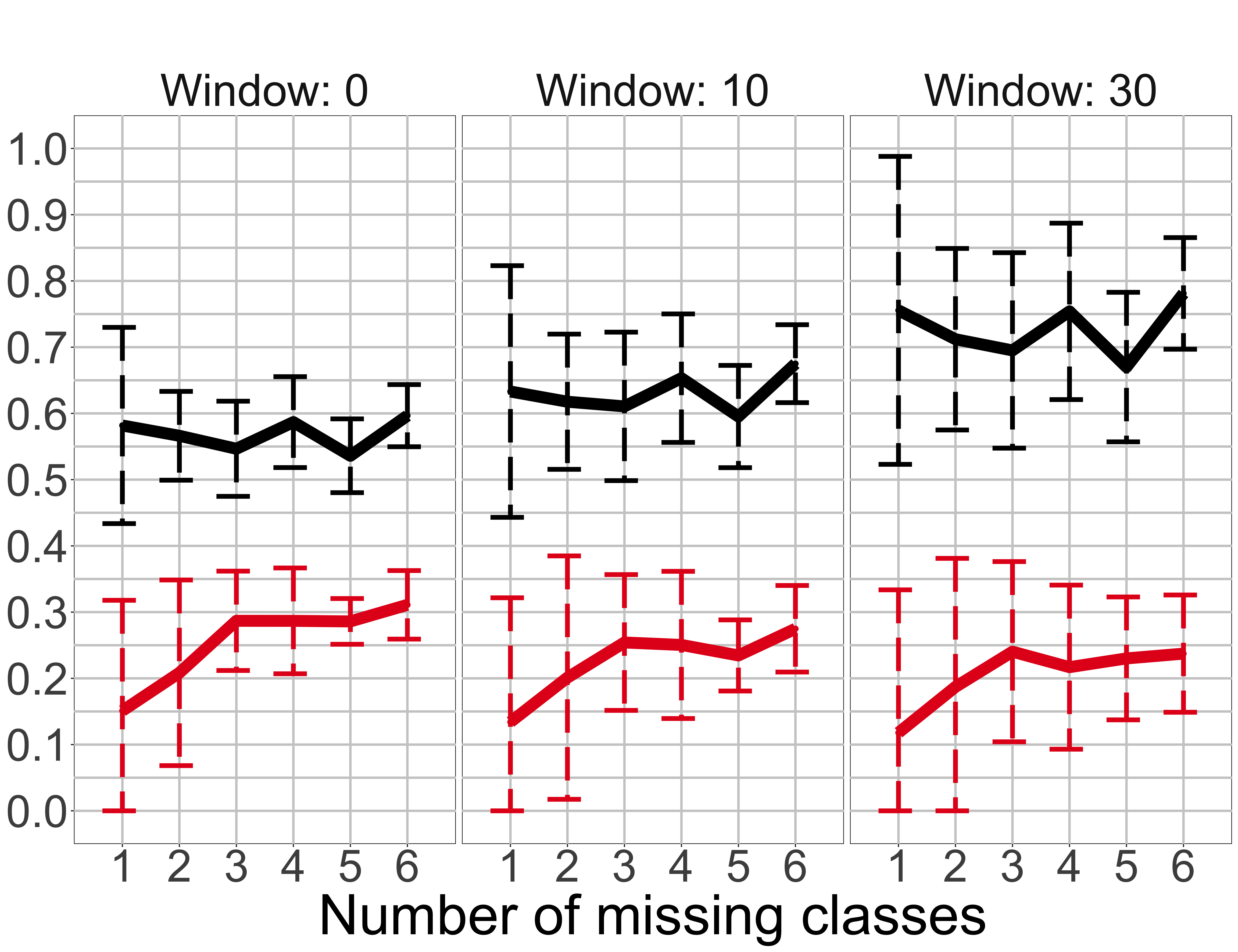} 
\caption{Nonconsecutive missing classes\label{fig:miss_noseq}}
\end{subfigure}
\begin{subfigure}{0.48\textwidth}
\centering
\includegraphics[trim={1cm 2cm 1cm 9.5cm}, clip, height=0.18\textheight]{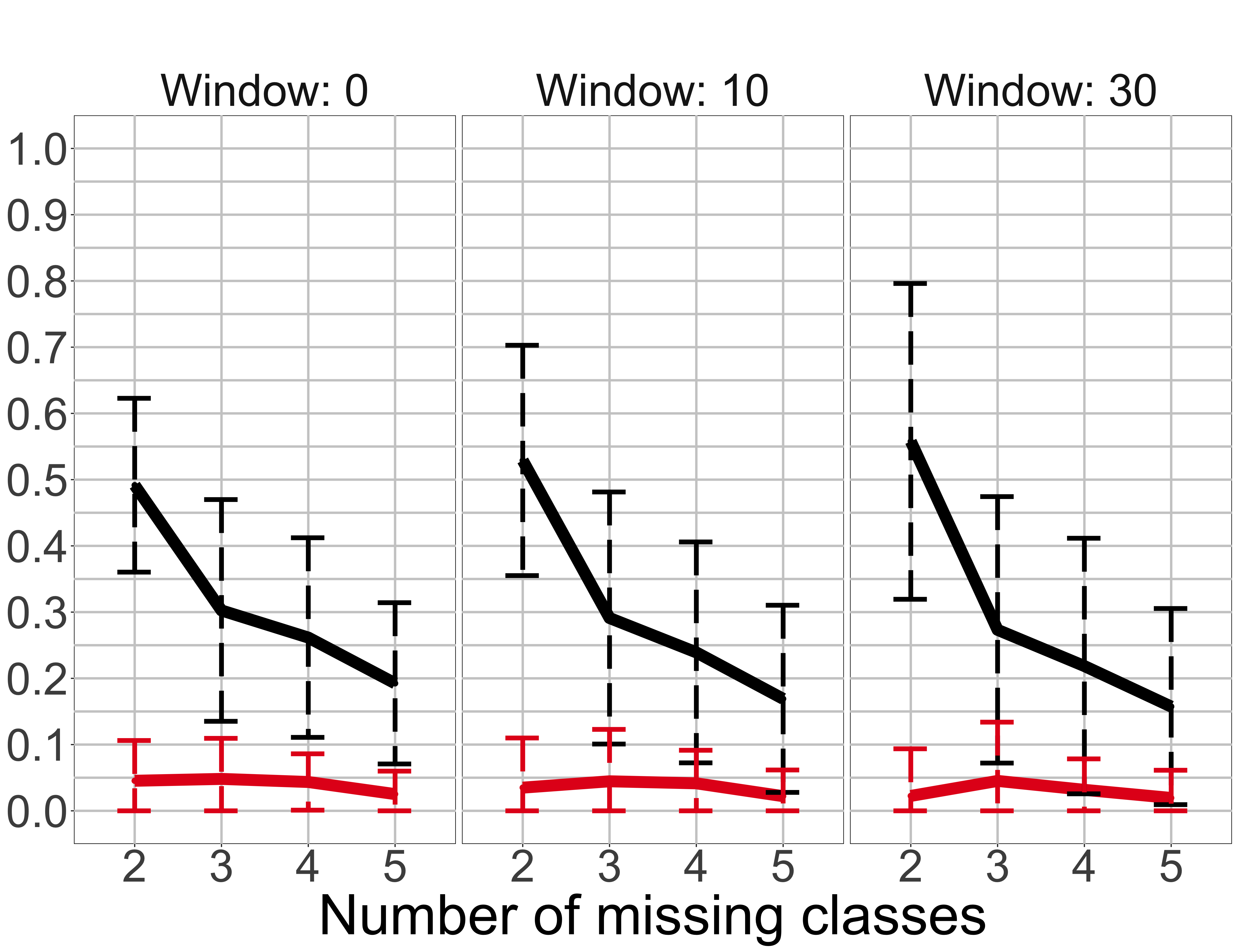} 
\caption{Consecutive missing classes\label{fig:miss_seq}}
\end{subfigure}

\caption{Accuracy of Methods: black is our (OQ + test + rank correlation), red is baseline (triplet + interpolation). We compare the average accuracy of missing classes (solid lines) and all classes (dashed lines) depending on: 1) types of missing classes: nonconsecutive or consecutive; 2) number of missing classes; 3) size of window correction. For any number of missing classes our method outperforms baseline. It is clear that correction mechanism improves accuracy with non-consecutive missing classes and average overall accuracy achieves the level of full data accuracy. 
\label{fig:miss}}
\end{figure*}

\begin{figure}[ht!]
\centering

\begin{subfigure}{0.2\textwidth}
\centering
\includegraphics[trim={4.2cm 5cm 4.2cm 3.6cm}, clip, height=0.16\textheight]{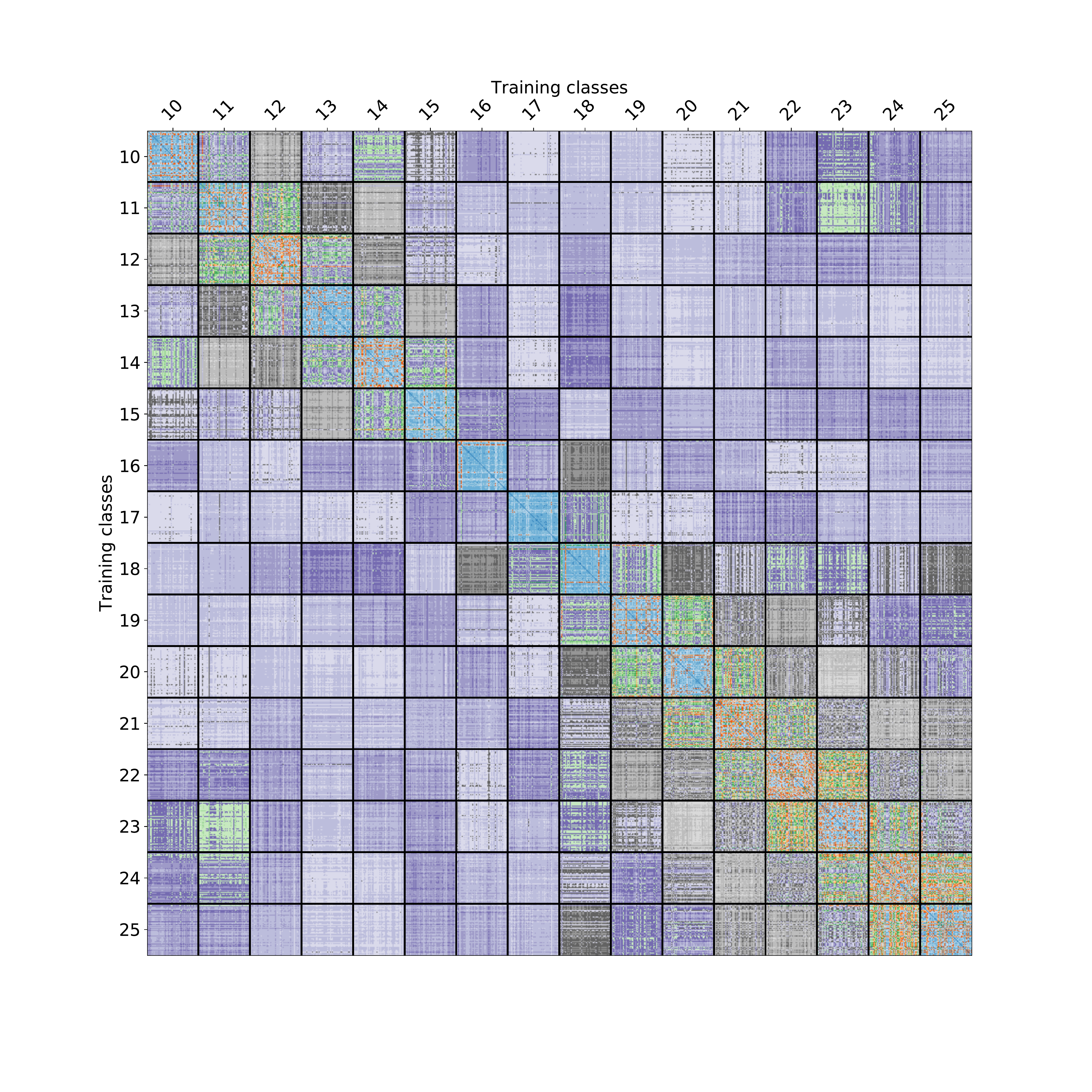} 
\caption{Triplet \label{fig:dist_triplet}}
\end{subfigure}
\begin{subfigure}{0.2\textwidth}
\centering
\includegraphics[trim={4.2cm 5cm 4.2cm 3.6cm}, clip, height=0.16\textheight]{figs/fds/fd_ordinal_quadruplet.pdf} 
\caption{Our (OQ) \label{fig:dist_oq}}
\end{subfigure}
\begin{subfigure}{0.05\textwidth}
\centering
\includegraphics[trim={31cm 4.95cm 5cm 4.6cm}, clip, height=0.16\textheight]{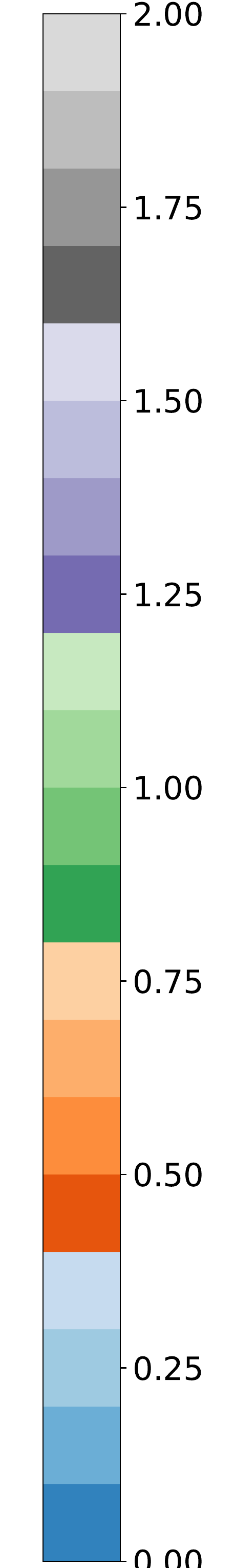} 
\caption*{}
\end{subfigure}

\caption{pairwise distances in feature space of the trained encoder between samples of different classes on Optical SNR dataset. Encoder is trained for time series classification problem with 16 classes. (a) Triplet loss does not preserve order of distances; (b) Our loss, ordinal-quadruplet with $D_y = |a-b|$, preserves order and force encoder to learn class similarity.   \label{fig:dist} }
\end{figure}

\section{Experiments}
We measure the effectiveness of the proposed framework using two criteria: 1) ability to preserve the label order into feature space; and 2) prediction performance. 
We evaluate our model using three real world time series data sets, and compare with the baseline, which is a triplet loss based optimization augmented with interpolation  when there are missing labels (as described in Section~\ref{subsec:statement}). 
In case when classes are missing from a training data, our model shows significantly better results than the baseline, with accuracy for missing classes higher in 3 times on average in some scenarios. Meanwhile, when both training and testing data have the same classes,  our method is at least as good as the baseline.

\begin{figure*}[ht!]
\centering

\begin{subfigure}{0.23\textwidth}
\centering
\includegraphics[trim={0.9cm 0cm 0.5cm 1.5cm}, clip, width=0.95\textwidth]{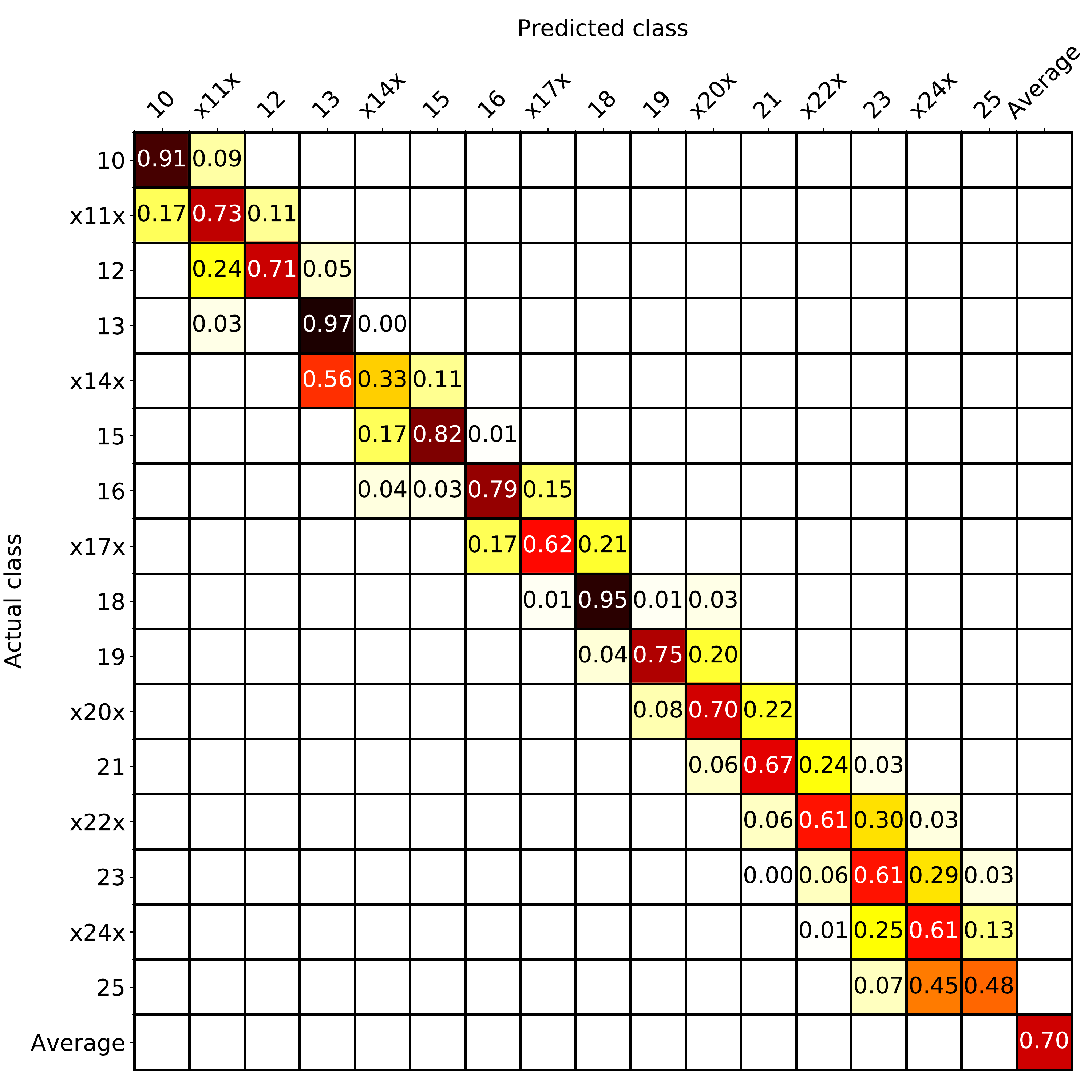}
\caption{Our, $w=0$ }
\end{subfigure}
\begin{subfigure}{0.23\textwidth}
\centering
\includegraphics[trim={0.9cm 0cm 0.5cm 1.4cm}, clip, width=0.95\textwidth]{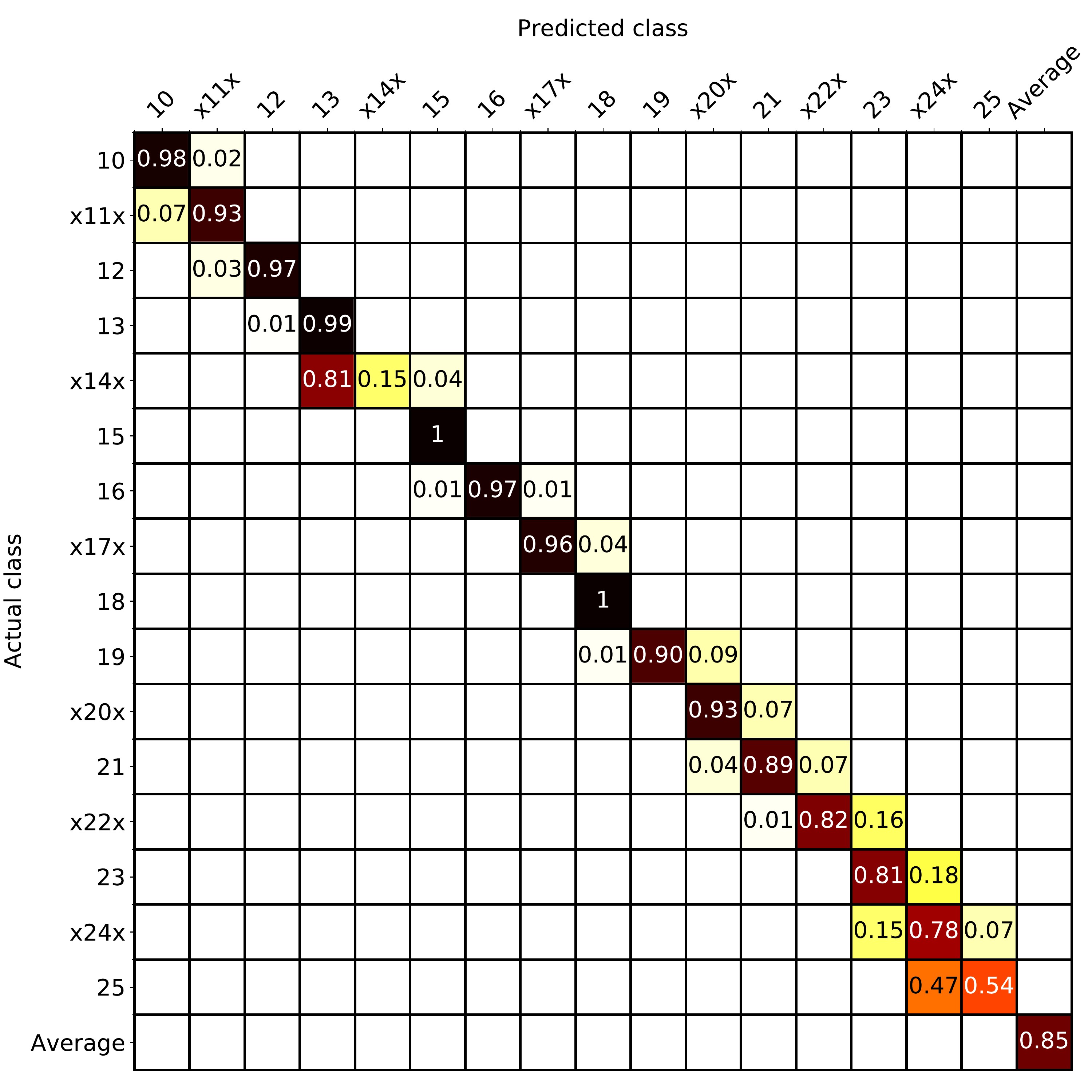}
\caption{Our, $w=30$}
\end{subfigure}
\begin{subfigure}{0.23\textwidth}
\centering
\includegraphics[trim={0.9cm 0cm 0.5cm 1.5cm}, clip, width=0.95\textwidth]{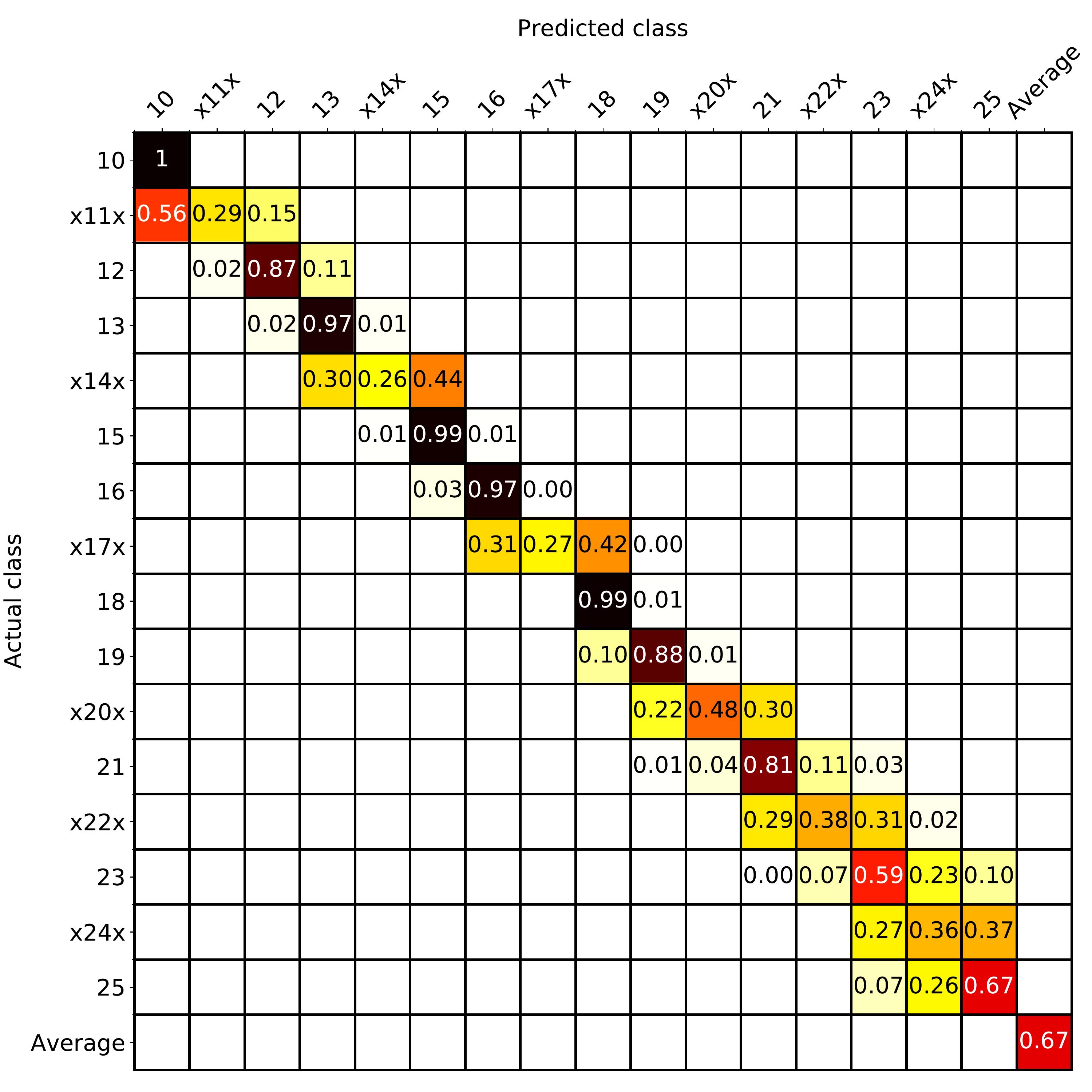}
\caption{Triplet, $w=0$}
\end{subfigure}
\begin{subfigure}{0.23\textwidth}
\centering
\includegraphics[trim={0.9cm 0cm 0.5cm 1.5cm}, clip, width=0.95\textwidth]{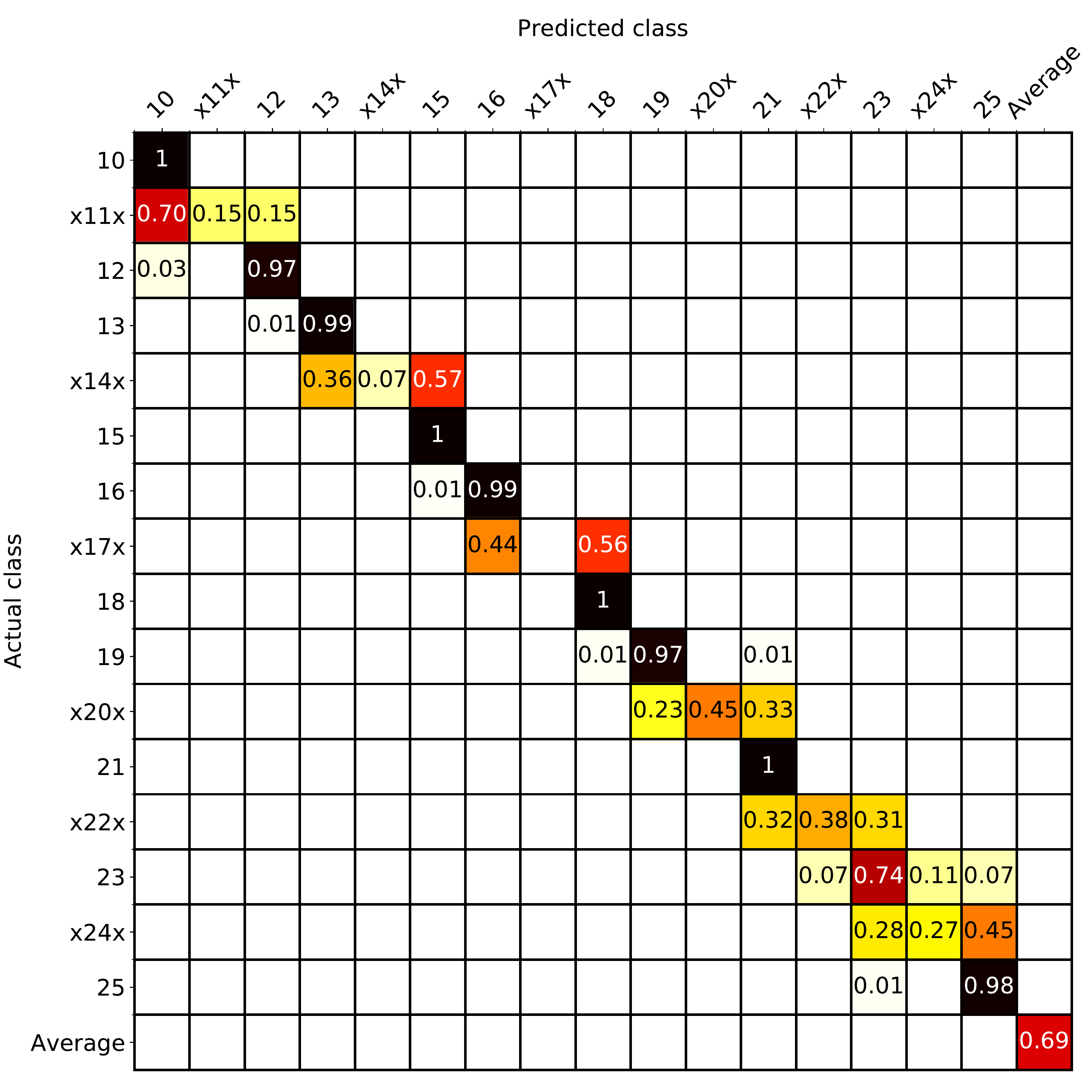}
\caption{Triplet, $w=30$  }
\end{subfigure}

\caption{Accuracy of prediction given missing classes (labels surrounded by x) in training data for two methods: (a, b) Our (OQ) loss and hypothesis testing, and (c, d) is Triplet and interpolation. Different values of $w$ indicates the size of correction window. Clearly correction window is beneficial for our method, because of the nature of hypothesis test and proposed loss. \label{fig:cm_random}}
\end{figure*}

\subsection{Datasets}

We use three time series datasets to demonstrate the performance of our platform, which represent two different scenarios where ordinal classification is useful: when labels are inherently ordered, \ie{} are numbers or comparative adjectives (as in the Optical SNR data) and when the order is not clearly defined and must be inferred or assumed based on characteristics of the data (as in the Audio data set) or general domain knowledge (as in the HAPT data). Our framework is oblivious to the actual order relationship; it simply seeks to transfer the label order into the embedding space. 

{\bf Optical SNR} data consists of 200 time series representing parameters of an optical transceiver. Each time series contains values recorded at every second for almost nine hours. The data is labeled with the measured SNR value of the light through the transceiver. There are 16 labels corresponding to SNR values from 10 to 25dB. These SNR values implicitly establish an order relationship between the labels.

{\bf Audio} data consists of brief (\ie{}, less than 10s) sound recordings from ten urban settings such as park, street, playground, or subway station~\cite{audio-data-source}. As in previous audio scene detection work~\cite{d2daudio}, we pre-process each recording to extract MFCC raw audio features~\cite{mfcc}. The pre-processed data contains 40 time series; each time series corresponds to the values of one MFCC band. We order labels according to the average loudness across their associated scenes, extracted using {\em pyloudnorm}~\cite{audio-loudness-tool}. The order matches our expectations; for example, we rank subway stations and intersections louder than playgrounds and parks.

{\bf HAPT} (Smartphone-Based Recognition of Human Activities and Postural Transitions Data Set \cite{reyes2016transition}) is an activity recognition data set built from the recordings of subjects performing basic activities and postural transitions, while wearing a waist-mounted smartphone with embedded inertial sensors. The 561 extracted features are recorded at every time step; they characterize one of six main labels: 
laying (1), sitting (2), standing (3), walking down (4), walking (5), walking up (6).  Since classes represent human physical activities, it can be assumed that there is an ordinal relation between \cite{feldman2003physical, satake2018sparse}.
We follow the order established by \cite{hsu2018cleaver} and indicated by the numbers between brackets.
   
\subsection{Parameter Settings}
To learn feature representation of time series data we use Bidirectional-RNN (Bi-RNN) encoder \cite{jagannatha2016bidirectional}. For all experiments we fix mini-batch as 256 and learning rate as 0.005. We set the hidden feature dimension of Bi-RNN as 256 and window size as 10. Finally, on the top of Bi-RNN we have a fully connected layer with 256 dimensions as an output. 
Following \cite{wang2017normface}, we use $L_2$ normalization on the resulted features.
Encoder was implemented in TensorFlow and all experiments were conducted on NVIDIA GeForce RTX 2080 Ti. 
For the our method, Ordinal Quadruplet loss, we choose the label distance function as $D_y(i, j) = |i -j|$. As a rank-based statistics we chose Kendall's $\tau$ version, which deals with ties. To control $P(\text{type I error})$, we set $\alpha = 0.05$. We discuss the choice of window size for historical window correction when we describe the experiment results.

\begin{figure}[ht!]
\centering
\includegraphics[trim={1cm 2.6cm 1cm 9.5cm}, clip, height=0.17\textheight]{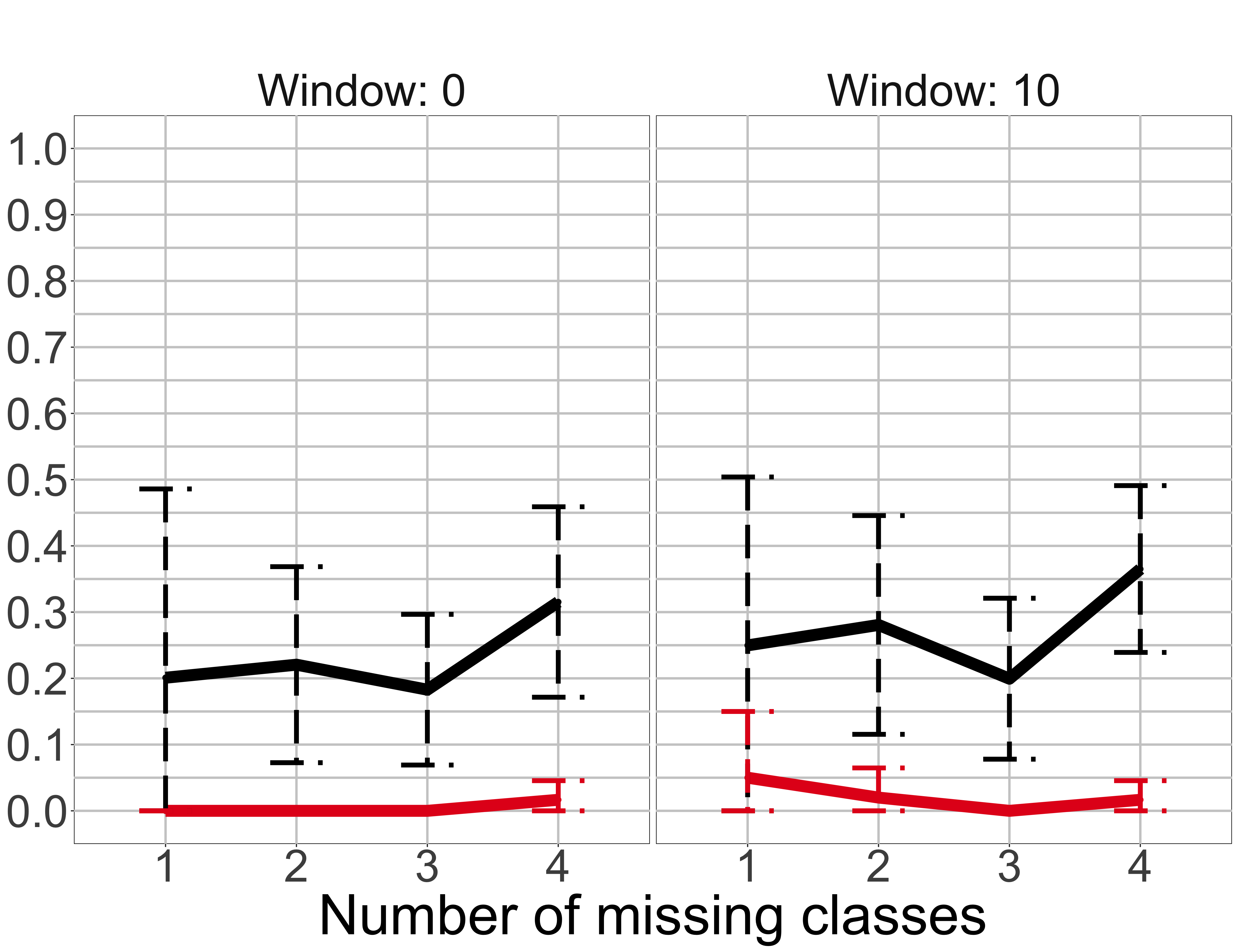} 
\caption{Accuracy of nonconsecutive missing classes in audio-scene detection. Methods: black is our (OQ + test + rank correlation), red is baseline (triplet + interpolation).\label{fig:miss_audio}}\vspace{-4mm}
\end{figure}

\begin{figure*}[ht!]
\centering
\begin{subfigure}{0.8\textwidth}
\centering
\includegraphics[trim={4.4cm 7.3cm 3.4cm 30cm}, clip,
height=0.03\textheight]{figs/dist_train_train_hapt_miss_random_1-6_random_0_triplet_1000.pdf} 
\end{subfigure}

\begin{subfigure}{0.3\textwidth}
\centering
\includegraphics[trim={8.8cm 13cm 8cm 3.6cm}, clip, height=0.23\textheight]{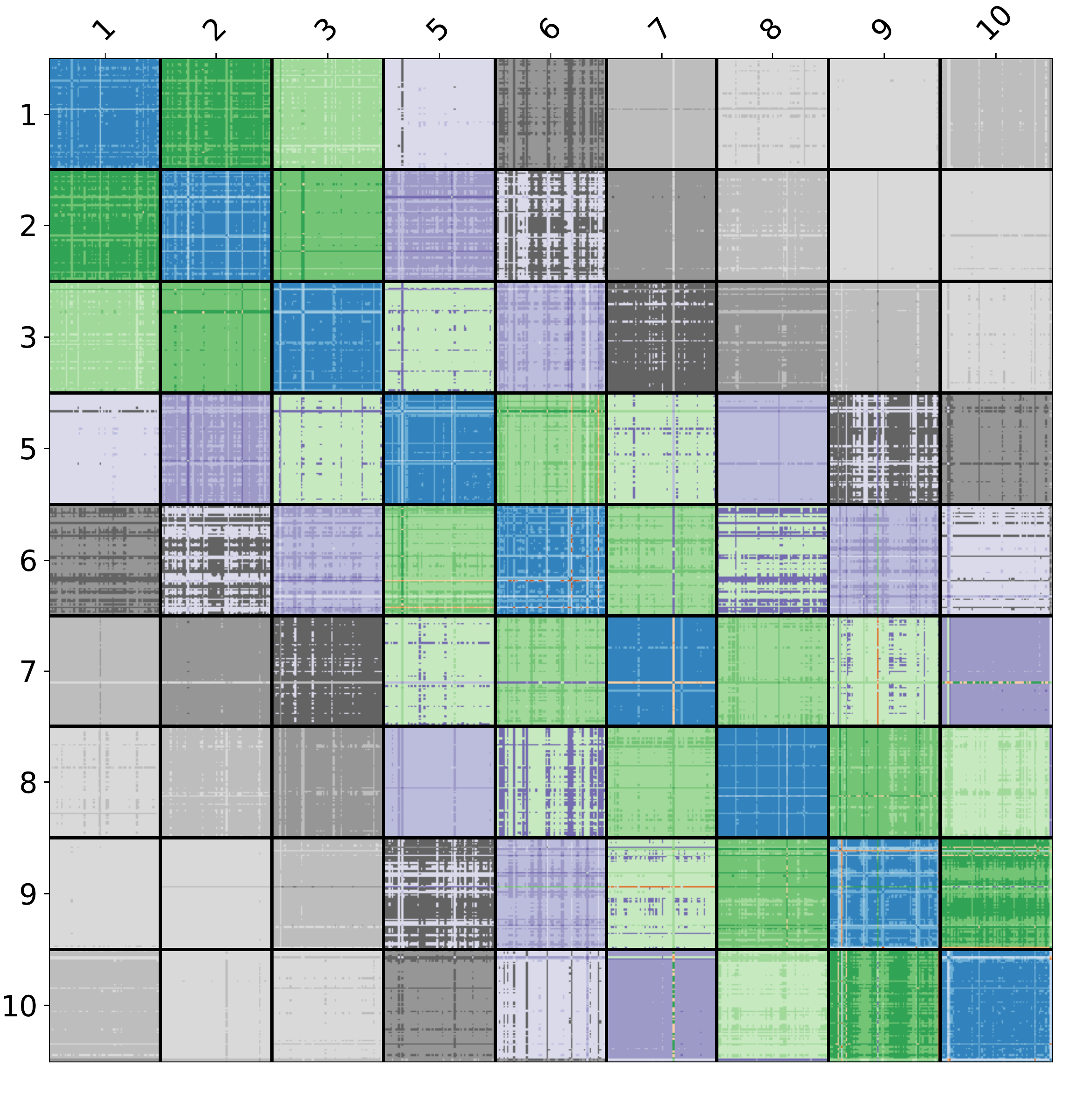}
\caption{Train}
\end{subfigure}
\begin{subfigure}{0.6\textwidth}
\centering
\includegraphics[trim={4cm 13cm 4cm 14.6cm}, clip, height=0.2\textheight]{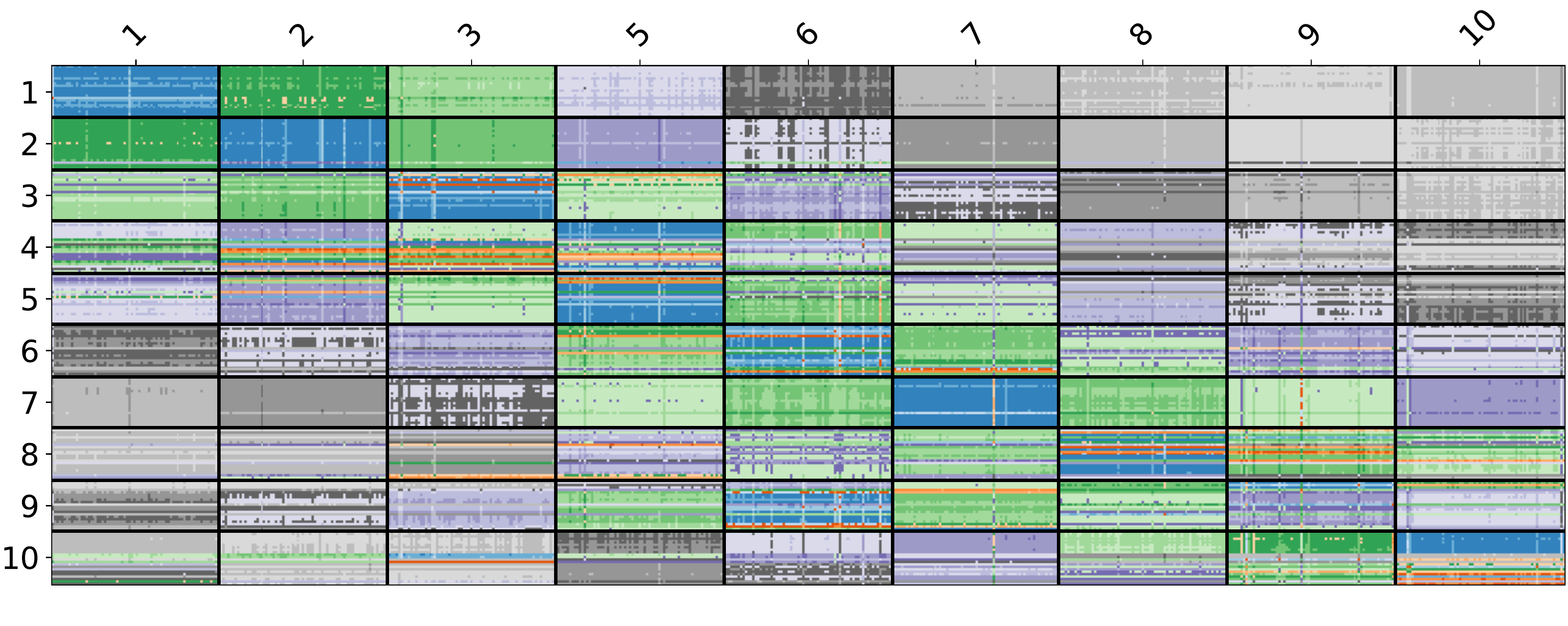}
\caption{Test}
\end{subfigure}
\caption{Pairwise distances in features space of the trained encoder between samples of different classes. For this experiment training data does not contain label "4". For testing set distance between learned features for missing label "4" located between "3" and "5" in respect of order of distances. Which makes it possible to apply hypothesis test predicting missing label.}
\label{fig:audio_order}
\end{figure*}

\begin{figure*}[ht!]
\centering
\begin{subfigure}{0.8\textwidth}
\centering
\includegraphics[trim={1cm 7.3cm 0 30cm}, clip,
height=0.03\textheight]{figs/dist_train_train_hapt_miss_random_1-6_random_0_triplet_1000.pdf} 
\end{subfigure}

\begin{subfigure}{0.45\textwidth}
\centering
\includegraphics[trim={3cm 13cm 3cm 13cm}, clip,
height=0.14\textheight]{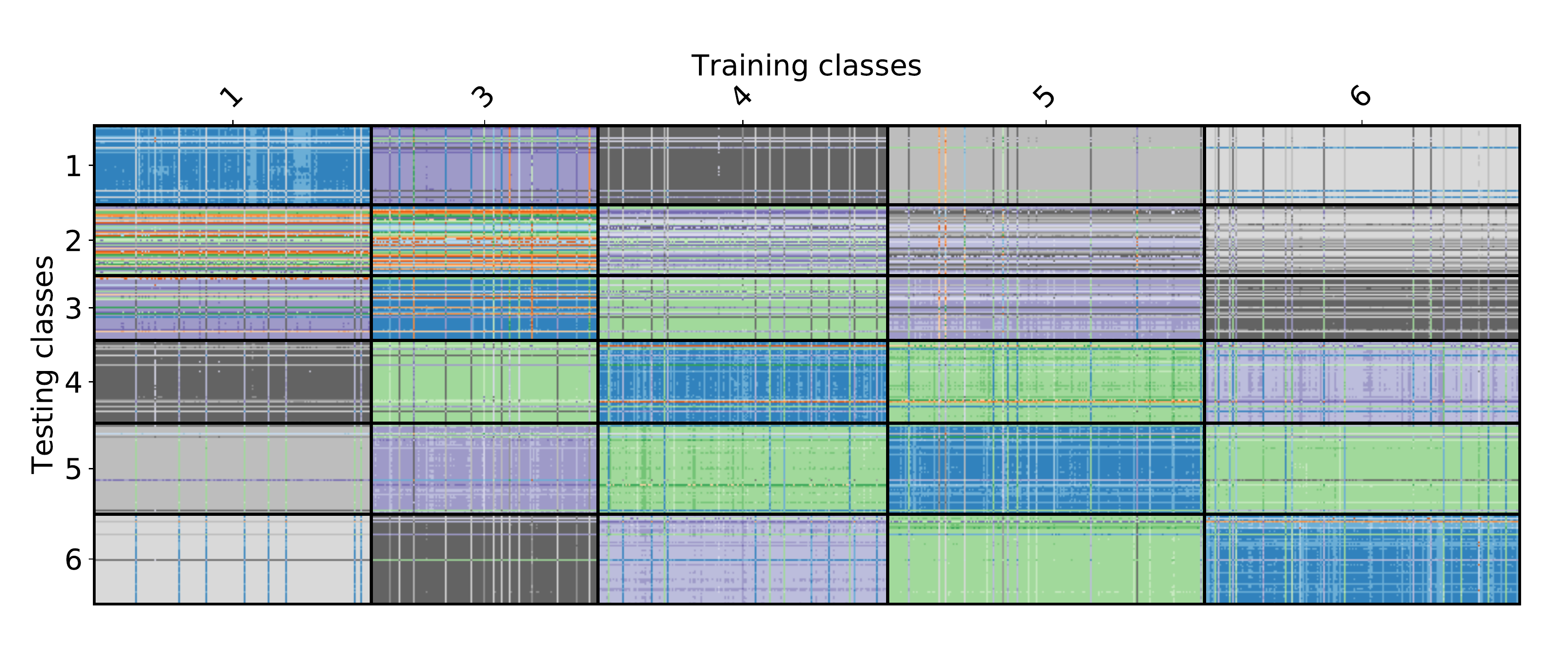} 
\caption{Our method: preserves order of distances in feature space and ``put" missing class between its neighbors in label space.\label{fig:dist_hapt_our}}
\end{subfigure}
\begin{subfigure}{0.45\textwidth}
\centering
\includegraphics[trim={3cm 13cm 3cm 13cm}, clip,
height=0.14\textheight]{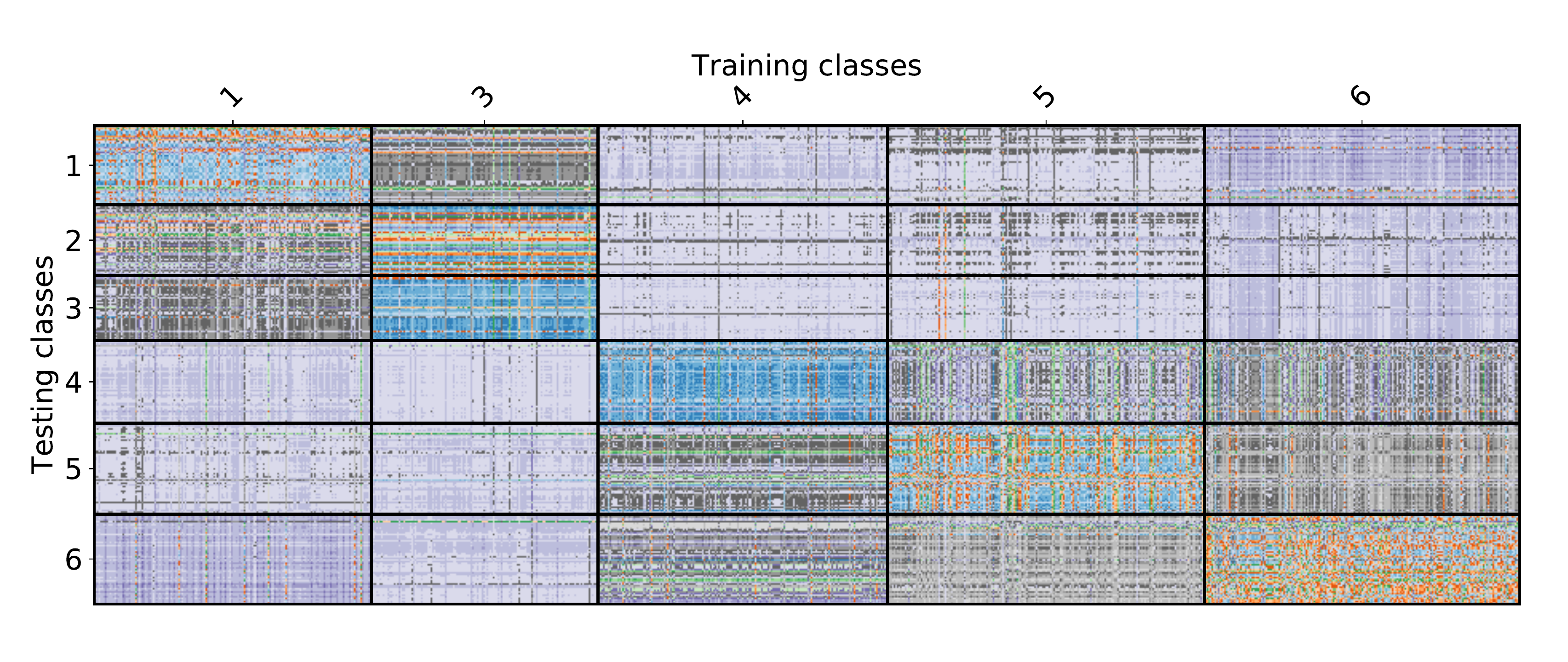}
\caption{Triplet loss: does not preserve order, but separates classes, with degradation in performance on classes 5 and 6.\label{fig:dist_hapt_triplet}}
\end{subfigure}

\caption{shows pairwise distances in feature space between samples of different classes. When distances are computed to features of training data (x-axis),  class 2 (sitting activity) is not available, while it is present in testing classes (y-axis).\label{fig:dist_hapt}}
\end{figure*}

\subsection{Experiments based on Optical SNR}
\label{subsec:osnr}
First, we demonstrate that our method preserves the order of distance in feature space, and shows improvements compare to other methods. Then, we discuss its performance in retrieving missing information during prediction. 

\textbf{Order preservation.} 
We compute features for time series data using two different optimization loss functions: triplet, and ordinal-quadruplet. We compute the pairwise Euclidean distance between all feature vectors and show it as a heatmap in Figure~\ref{fig:dist}. 
Compared to triplet loss (Figure~\ref{fig:dist_triplet}, ordinal-quadruplet (Figure~\ref{fig:dist_oq}) preserves the order of the label space (compare the color bar with the progression of colors from the diagonal to the corners) and learns class similarity, which results in smaller distances between features of samples in the same class (blue squares on the diagonal).

\textbf{Retrieval of missing information.}
Since SNR data contains wide range of ordinal labels, to evaluate the power of our framework to predict classes missing from the training data,  we conduct experiments with two different types of missing classes:
\begin{enumerate}
    \item nonconsecutive, \ie{}, at least one non-missing class between missing classes. We consider up to six missing classes, which is 40\% of all classes in the training set. Since for every number of missing classes we can find a set of different classes to disregard, we evaluate performance on 14 different sets.
    \item consecutive, \ie{}, missing classes follow each other. We consider up to five missing classes in a row. For every number of missing classes $N_\text{miss}$, we choose 10 random starting points and select the following $N_\text{miss}-1$ classes as a set to disregard.
\end{enumerate} 
Conducting experiments for several $N_\text{miss}$ values allows us to evaluate the accuracy based on confidence intervals (CI) and not rely on a point estimate.  In addition to different types of missing classes, we evaluate the effect of window size for historical correction. We compare our method with triplet loss combined with interpolation.

Figure~\ref{fig:miss_noseq} shows the accuracy of predicting samples from missing classes when missing classes are \textbf{nonconsecutive} and we vary their number. Each of the three plots corresponds to a different window size when doing window correction (size 0 means we are not doing any correction). Our method is almost twice as accurate as baseline. We see that our framework benefits from window correction technique by improving type I error and increasing accuracy for missing classes, from around 0.5 to almost 0.7, meanwhile it does not affect the baseline method.
 
Figure \ref{fig:miss_seq} shows results for \textbf{consecutive} missing classes. With two missing classes our method significantly outperforms interpolation for missing class detection (0.35 vs 0.05 accuracy). As expected, with increasing number of consecutive missing classes it is harder to preserve order, which results in performance drop for both methods. However, our method still provides higher results. 
 
\textbf{Visualization of 1 experiment.} While it is clear that our method outperforms the baseline for any number of nonconsecutive missing classes, in Figure \ref{fig:cm_random} we show the prediction accuracy in details for one (out of 14) experiment, where data of 6 out of 13 classes (46\%) was not available during the training. Even when we do not perform window correction (\ie{} $w=0$), our method gives a higher accuracy (averaged through all classes), and much better prediction for missing classes, \eg{} 11, 17, 20, 22, and 24. With increasing length of correction window $w$, we achieve 0.88 average accuracies on all classes, compared to 0.67 achieved by the baseline with no window correction, and 0.75 with window correction.

\begin{figure*}[ht!]
\centering
\begin{subfigure}{0.18\textwidth}
\centering
\includegraphics[trim={1cm 0 0 1.2cm}, clip,
width=\textwidth]{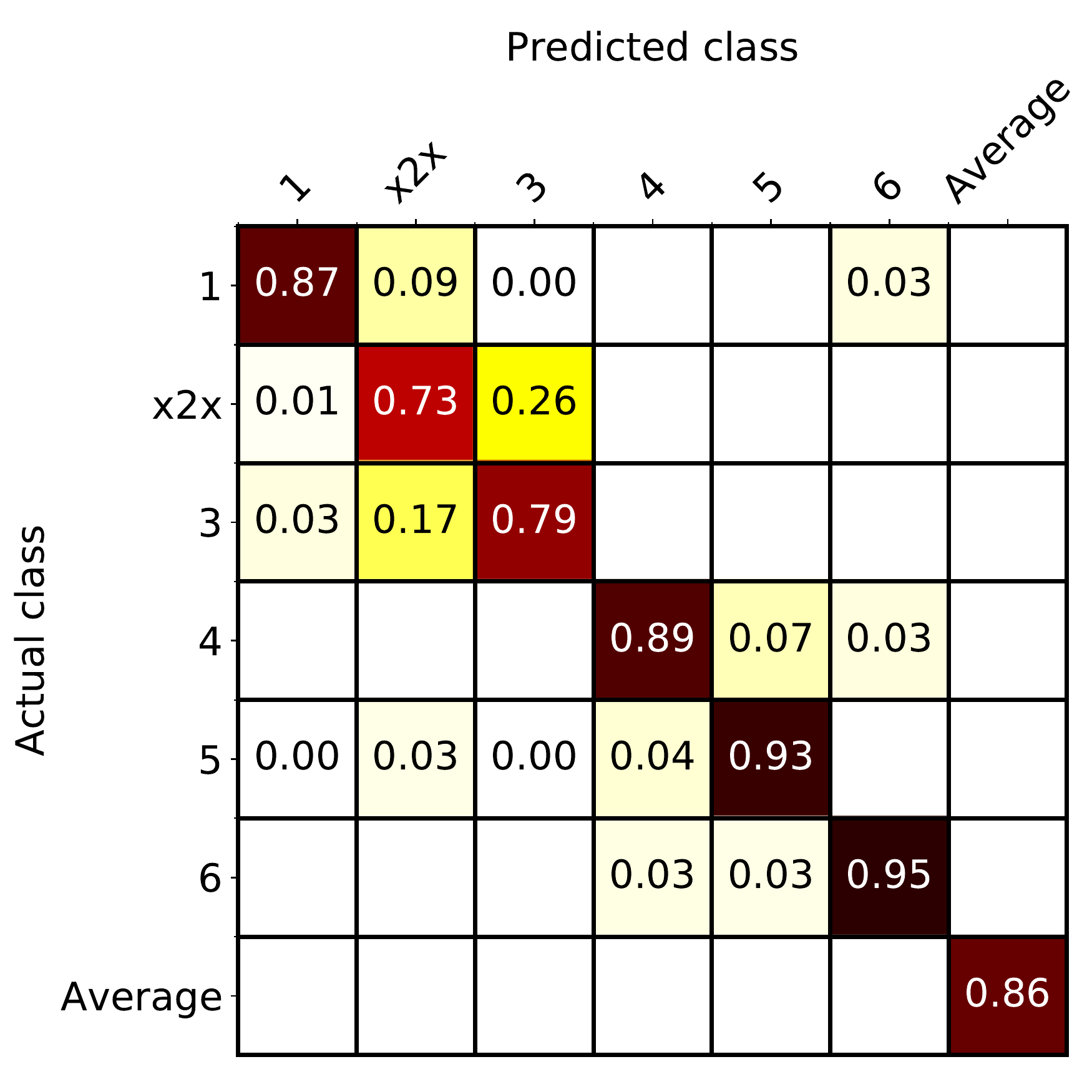} 
\caption{Our method with\\$\alpha=0.05$ and $w=0$\label{fig:cm_hapt_our}}
\end{subfigure}
\begin{subfigure}{0.18\textwidth}
\centering
\includegraphics[trim={1cm 0 0 1.2cm}, clip,
width=\textwidth]{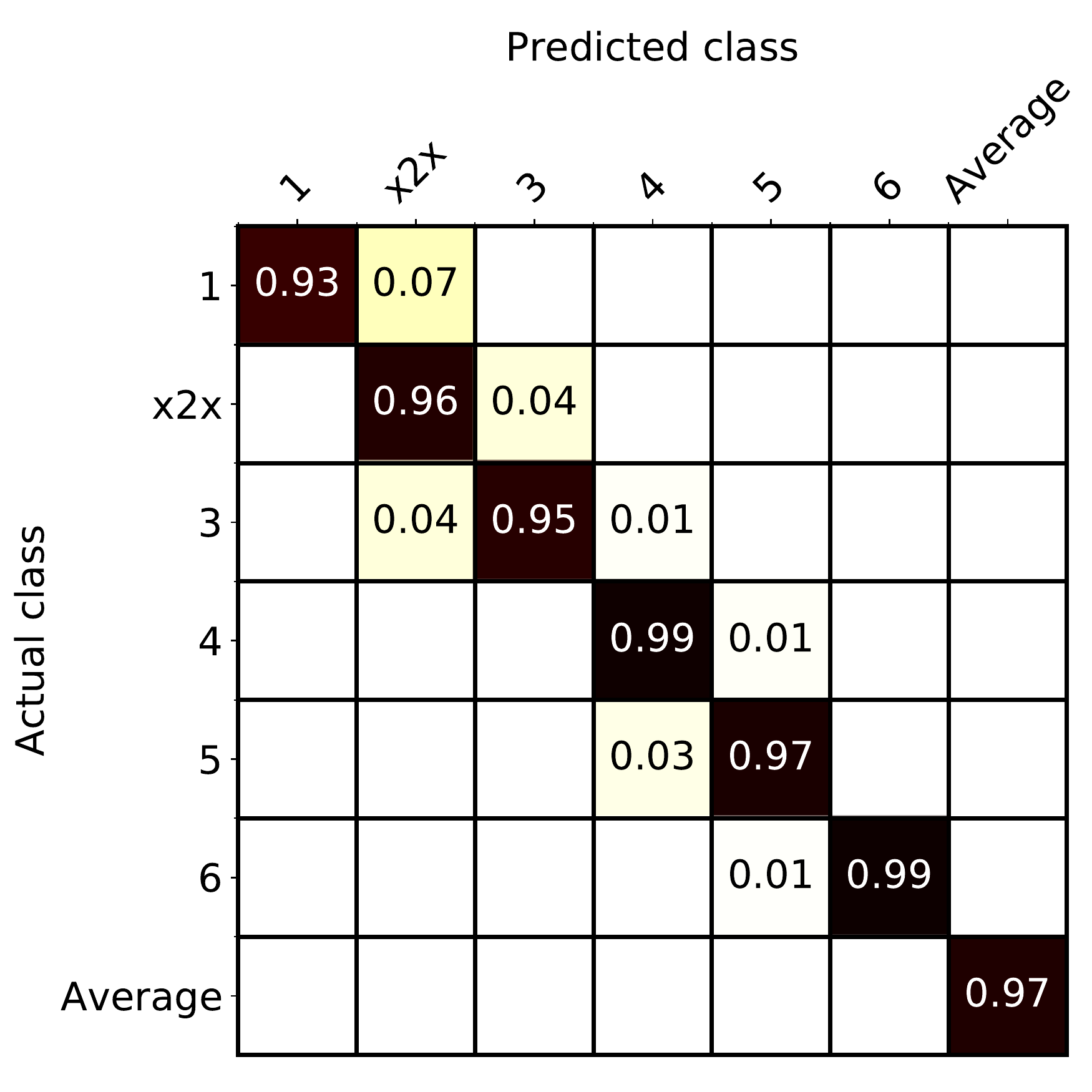} 
\caption{Our method with\\$\alpha=0.05$ and $w=10$\label{fig:cm_hapt_our_w}}
\end{subfigure}
\begin{subfigure}{0.18\textwidth}
\centering
\includegraphics[trim={1cm 0 0 1.2cm}, clip,
width=\textwidth]{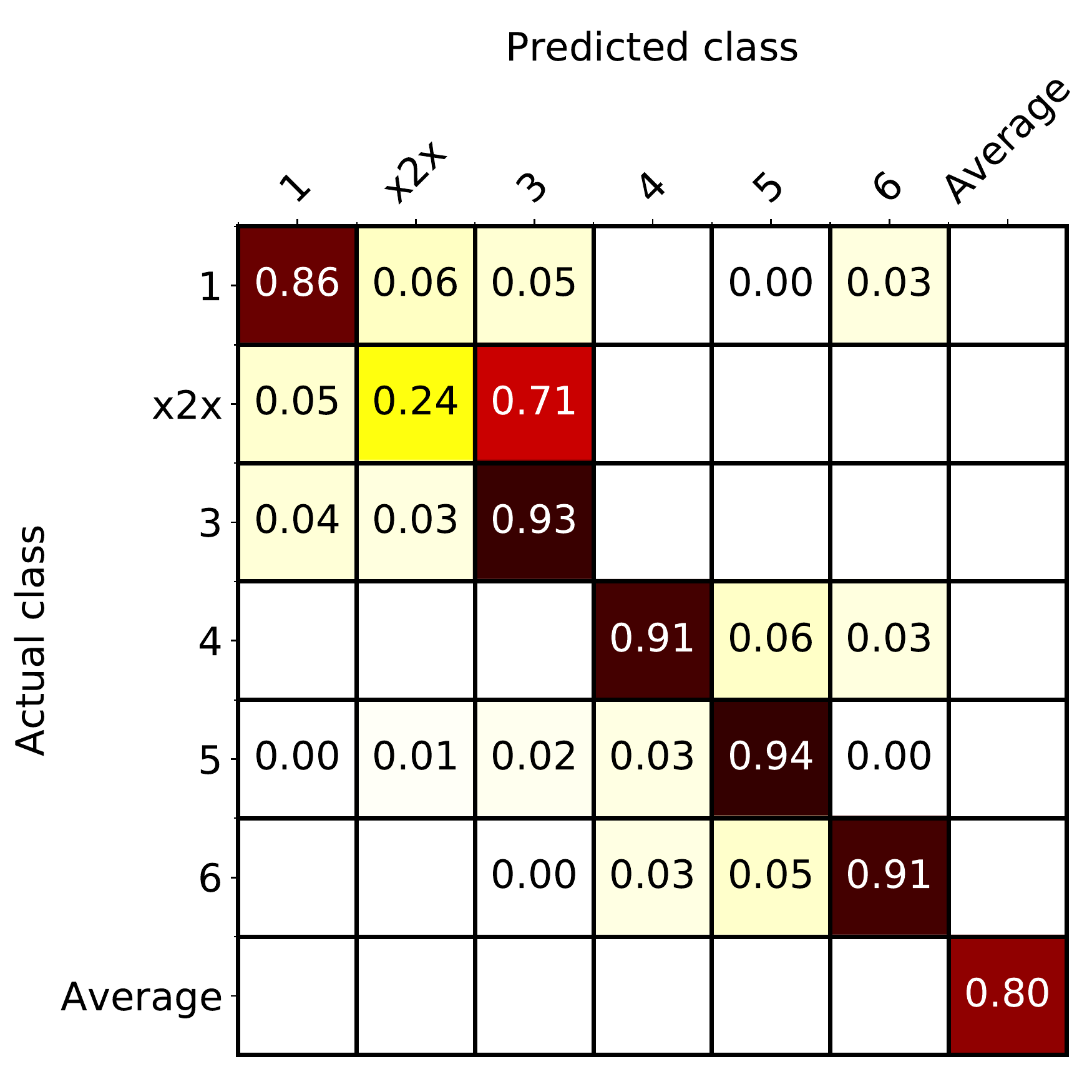}
\caption{Triplet loss with interpolation and $w=0$\label{fig:cm_hapt_triplet}}
\end{subfigure}
\begin{subfigure}{0.18\textwidth}
\centering
\includegraphics[trim={1cm 0 0 1.2cm}, clip,
width=\textwidth]{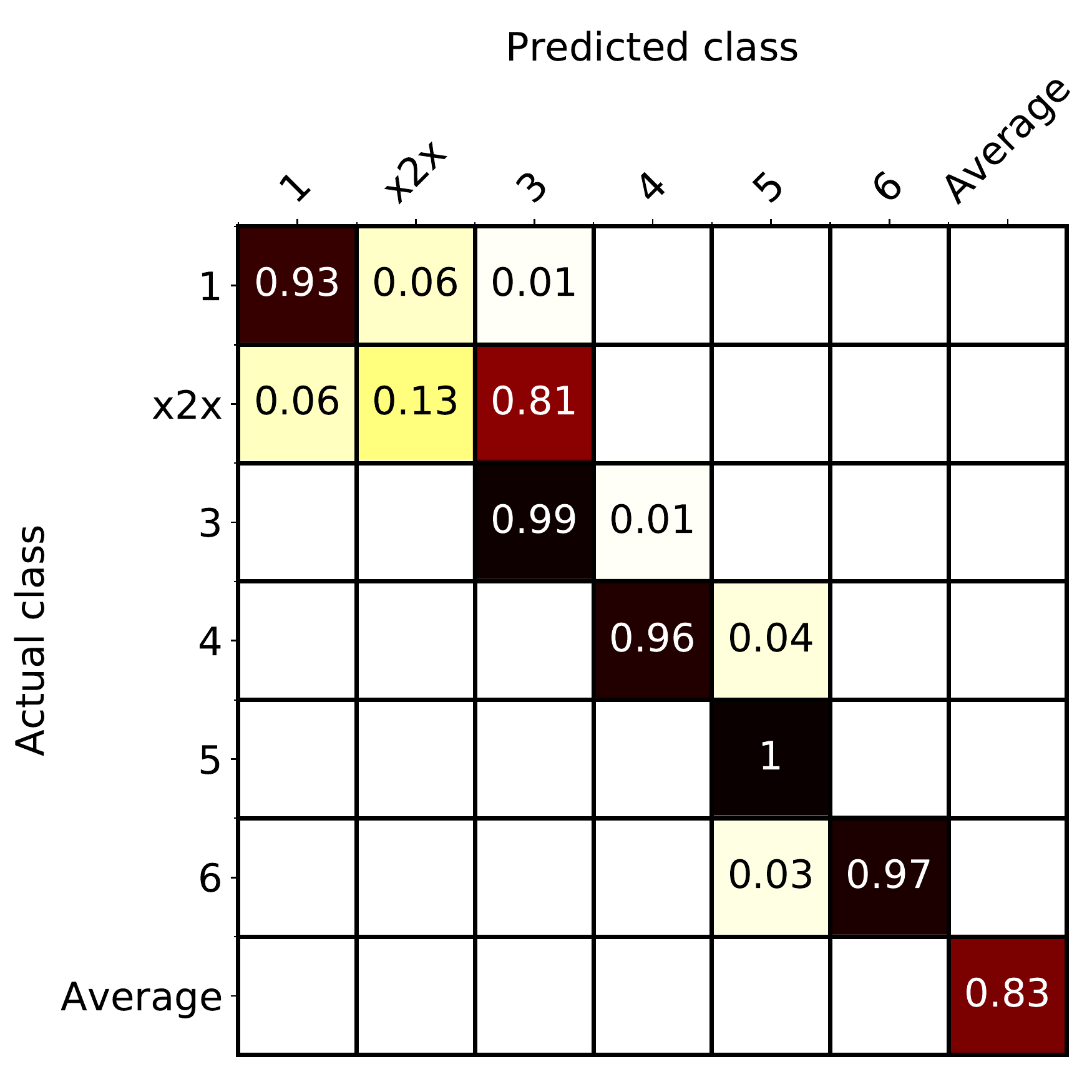} 
\caption{Triplet loss with interpolation and $w=10$\label{fig:cm_hapt_triplet_w}}
\end{subfigure}
\begin{subfigure}{0.18\textwidth}
\centering
\includegraphics[trim={1cm 0 0 1.2cm}, clip,
width=\textwidth]{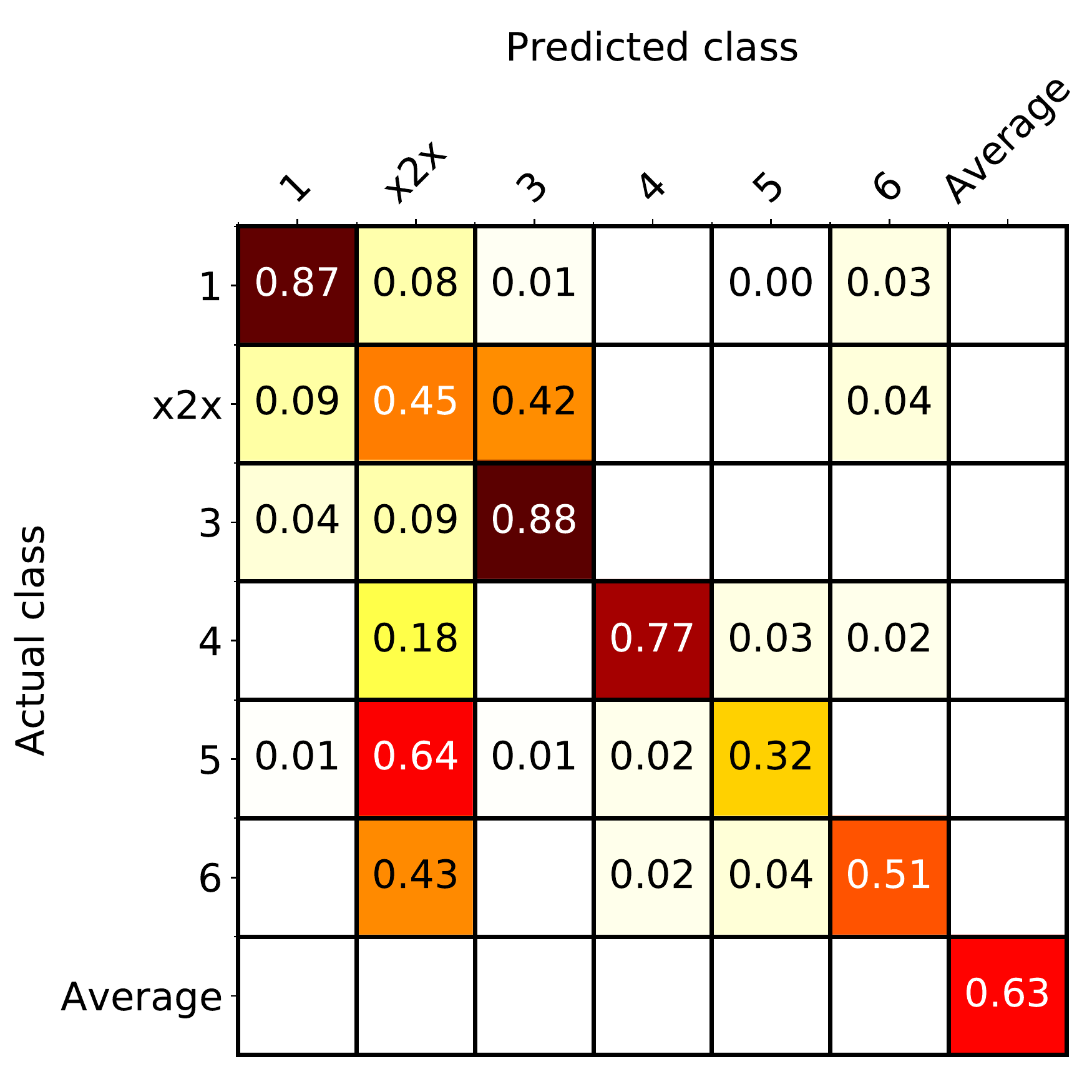} 
\caption{Triplet loss with $\alpha=0.05$ and $w=0$\label{fig:cm_hapt_triplet_test}}
\end{subfigure}
\caption{All: x-axis is predicted class; y-axis is real class; x2x is missing class. Classes (activities): 1 - laying, 2 - sitting, 3 - standing, 4 - walking downstairs, 5 - walking, 6 - walking upstairs. No correction: our method (a) predicts missing class with 73\% of accuracy, while triplet loss with interpolation (c) has accuracy of just 24\%. With correction: our method (b) get the highest average accuracy of 97\% and 96\% for missing class, while for interpolation (d) correction decrease performance for missing class. We see that our proposed hypothesis goes together with our proposed OQ loss and cannot be used with triplet loss, because triplet does not preserve order of distance. \label{fig:cm_hapt}}
\end{figure*}

\subsection{Experiments based on audio scenes}
We repeat the experiments in previous subsection and attempt to predict missing classes in the audio data. Figure~\ref{fig:miss_audio} presents the accuracy of missing classes when up to four nonconsecutive classes (or 40\% of the data) are missing from training.
Our method improves significantly upon the baseline: whereas the combination of triplet loss and interpolation almost never predicts the missing class correctly, we are able to achieve an accuracy of up to 50\%. 
Similar to SNR data we preserves order of distances in learned feature space. For example, Figure~\ref{fig:audio_order} shows the ability to preserve order in feature space, with missing classes for training and testing set. In this experiment class "4" was missing from the training.

Albeit much better than the baseline, our performance on the audio data is not as good as on the optical SNR data. We believe this happens because different audio scenes may contain similar segments (\eg{}, people talking both in parks and playgrounds), making it more difficult to compute representations that are far away from each other and thus more discriminative. 
We \textbf{emphasize} that our goal is to explore several real use cases for predicting missing labels given an order among all labels. As we have not yet customized our learning to take advantage of the structure of audio time series~\cite{d2daudio,spatiotemp,deeprnn,guo}, our results represent a worst case scenario. 


\subsection{Experiments based on HAPT}

For this experiment we explicitly remove the activity ``sitting" (class ``2") from the training.

\textbf{Order preservation.} 
Figure \ref{fig:dist_hapt_our} shows the feature distance heatmap for all non-missing classes. The order of labels is preserved in the feature space. In addition, distances to samples in the missing class ``2" are located (wrt to the color bar) between distances to classes ``1" and ``3", \ie{}, the missing activity ``sitting" is between activities ``laying" and ``standing". This means that during prediction, computing the two most similar classes based on distance ranking similarity will result in the most correlated labels as (``1", ``2") or (``3", ``2"). This allows us to perform hypothesis test. In the right side of Figure \ref{fig:dist_hapt_our}, we notice that the triplet loss does not preserve order well, which makes hypothesis test is not applicable.

\textbf{Prediction accuracy.} Figure \ref{fig:cm_hapt} shows the confusion matrices for classifying testing samples. This helps us understand 1) if our framework performs better than triplet + interpolation for retrieval of missing information; 2) if there is a benefit of window correction for hypothesis test and interpolation; and 3) if triplet loss can benefit from the test.

From Figures \ref{fig:cm_hapt_our} and \ref{fig:cm_hapt_triplet}, we observe that our method predicts missing class with 0.73 of accuracy, which is 0.5 higher than the baseline. Figures \ref{fig:cm_hapt_our_w} and \ref{fig:cm_hapt_triplet_w} show that our benefit persists even with window correction of size 10. Window correction increases the prediction for the missing class ``2'' from 0.73 to 0.96 and the overall average accuracy for all classes from 0.86 to 0.97. 

One may wonder why we do not use the hypothesis test, rather than interpolation, with the triplet loss. Figure~\ref{fig:cm_hapt_triplet_test} provides an answer to this question. 
Because triplet loss does not preserve order, it cannot select the most similar classes for the testing sample, leading to a lower accuracy of 0.45 for class ``2''. Even thous this is better than using interpolation, the hypothesis test performs poorly on the non-missing classes. Indeed, the overall accuracy when using hypothesis test instead of interpolation, drops from 0.8 to 0.63.

\section{Conclusions}
We proposed a novel framework for classifying ordinal time series data. Two critical components ensure the performance of our framework: an ordinal-quadruplet based optimization that forces the model to learn latent representation for time series, preserving the order of the label between classes; and a hypothesis testing based retrieval that identifies the correct label for each new data, even when data of that class is completely missing from the training.

Our framework,
has applications beyond classifying data from missing labels,
which brings \textbf{new directions} for research.
It can help domain adaptation by identifying new states as the data changes: as long as there is an order relationship between the states, our framework could identify the ``location'' of the new states in the embedding space, without knowing even one sample from the new state. Furthermore, the order can help to explain analysis results by identifying the time series that is most responsible for preserving the order.


         



\bibliography{refs}

\end{document}